\documentclass[acmsmall]{acmart}

\AtBeginDocument{%
  }

\setcopyright{acmcopyright}
\copyrightyear{2023}
\acmYear{2023}
\acmDOI{XXXXXXX.XXXXXXX}

\acmJournal{TOMM}
\acmVolume{37}
\acmNumber{4}
\acmArticle{111}
\acmMonth{8}

\usepackage{graphicx} 
\usepackage{multirow} 
\usepackage{tabularx} 
\usepackage{booktabs}
\usepackage{url}

\usepackage{hyperref}
\usepackage{multicol} 
\usepackage{subfigure}
\usepackage{caption}
\usepackage{float}

\usepackage{hyperref}
\usepackage{url}
\usepackage{graphicx} 
\usepackage{multirow} 
\usepackage{tabularx} 
\usepackage{booktabs}
\usepackage{url}
\usepackage{multicol} 
\usepackage{graphicx}
\usepackage{subfigure}
\usepackage{caption}
\usepackage{float}
\usepackage{color} 
\usepackage{setspace}

\begin{document}

\title{Sentiment-oriented Transformer-based Variational Autoencoder Network for Live Video Commenting}


\author{Fengyi Fu}
\affiliation{%
  \institution{University of Science and Technology of China}
  \city{Hefei}
  \state{Anhui}
  \country{China}
  \postcode{43017-6221}
}
\email{ff142536f@mail.ustc.edu.cn}
\orcid{0000-0002-4870-8804}

\author{Shancheng Fang$^{*}$} %
\thanks{$^{*}$Corresponding authors. This work is supported by the  National Science Fund for Excellent Young Scholars under Grant No.62222212, and the National Natural Science Foundation of China under Grant, No.62102384, No.62302474.}
\affiliation{%
  \institution{University of Science and Technology of China}
  \city{Beijing}
  \country{China}}
\email{fangsc@ustc.edu.cn}

\author{Weidong Chen}
\affiliation{%
 \institution{University of Science and Technology of China}
 \city{Hefei}
 \state{Anhui}
 \country{China}}
\email{chenweidong@ustc.edu.cn}

\author{Zhendong Mao}
\affiliation{%
 \institution{University of Science and Technology of China}
 \city{Hefei}
 \state{Anhui}
 \country{China}}
\email{zdmao@ustc.edu.cn}

\renewcommand{\shortauthors}{F. Fu et al.}


\begin{abstract}
Automatic live video commenting is with increasing attention due to its significance in narration generation, topic explanation, etc. However, the diverse sentiment consideration of the generated comments is missing from the current methods. Sentimental factors are critical in interactive commenting, and lack of research so far. 
Thus, in this paper, we propose a Sentiment-oriented Transformer-based Variational Autoencoder (So-TVAE) network which consists of a sentiment-oriented diversity encoder module and a batch attention module, to achieve diverse video commenting with multiple sentiments and multiple semantics. Specifically, our sentiment-oriented diversity encoder elegantly combines VAE and random mask mechanism to achieve semantic diversity under sentiment guidance, which is then fused with cross-modal features to generate live video comments. Furthermore, a batch attention module is also proposed in this paper to alleviate the problem of missing sentimental samples, caused by the data imbalance, which is common in live videos as the popularity of videos varies. Extensive experiments on Livebot and VideoIC datasets demonstrate that the proposed So-TVAE outperforms the state-of-the-art methods in terms of the quality and diversity of generated comments. Related code is available at \textcolor[rgb]{0,0,1}{\href{https://github.com/fufy1024/So-TVAE}{\textcolor[rgb]{0,0,1}{https://github.com/fufy1024/So-TVAE}}}.
\end{abstract}



\received{20 February 2007}
\received[revised]{12 March 2009}
\received[accepted]{5 June 2009}

\maketitle

\section{Introduction}
Live video commenting, commonly known as ``danmaku" or ``bullet screen", is a new interactive mode on online video websites that allows viewers to write real-time comments to interact with others. Recently, the automatic live video commenting (ALVC) task, which aims to generate real-time video comments for viewers, is increasingly used for narration generation, topic explanation, and video science popularization as it can assist in attracting the attention and discussion of viewers.

  \begin{figure}[!t]
  \vspace{-0.4cm}  
  \setlength{\abovecaptionskip}{0.cm} 
  \setlength{\belowcaptionskip}{-0.5cm}
\centering
\includegraphics[width=5.2in]{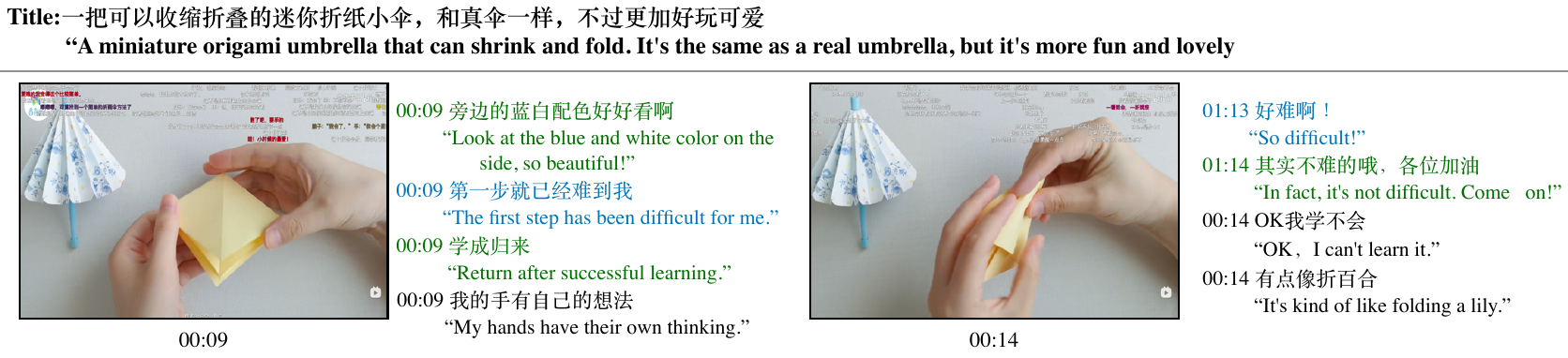}
\caption{A live video commenting example in Livebot with selected video frames and live comments. Green: Positive comments. Black: neutral comments. Blue: Negative comments.}
\label{fig_1}
\end{figure}

The preceding works~\cite{2018LiveBot,2020VideoIC,duan2020multimodal} simply transfer the methods from similar tasks (\textit{e.g.},~video captioning ~\citep{2020Learning,2016Sequence,xu2015show} ) to the ALVC task, which typically follow an encoder-decoder paradigm that first converts the input to a \emph{fixed} latent vector and thus reconstructs it into natural language text. Despite success in context-aware commenting, these traditional encode-decode frameworks intrinsically perform a one-to-one mapping, resulting in single and objective comments.

However, the ALVC is a more challenging task with the diverse and subjective characteristics of the comments. Figure \ref{fig_1} presents a real-life chatting scenario of video danmaku. Although the viewers discuss a common theme: origami umbrella, different viewers may make diverse comments from individual perspectives and sentiments, even at a single video frame. 
On the one hand, in real-world applications, it is difficult for comments with a single sentiment to resonate with everyone ~\citep{2021Distributed,2021Position}. On the other hand, diverse sentiment-guided comments would help video labeling and distribution, and further encourage human-interacted comment generation.
  Thus, in this paper, we first treat the ALVC task as a one-to-many generation task with sentimental distinction, to achieve diverse video commenting with multiple sentiments and multiple semantics.

Two major difficulties make this work extremely challenging. Firstly, sentiment, comment, and video are heterogeneous modalities, and they are both important for diverse video commenting. 
Previous works take their effort on sentiment-guided text generation ~\citep{2021Contrast,2022Emp,2021CEM} or video comment generation ~\citep{2018LiveBot,2020VideoIC}, but cannot apply to the complex situation that needs considering all three modalities simultaneously. 
Secondly, the imbalance of video data ~\citep{2021Knowing} hinders the generation of comments with the desired sentiment.
The imbalance in video data lies in two aspects: the popularity of videos and the sentimental tendency of videos, which together result in a lack of comment samples of certain sentimental types in some videos.

To this end, in this paper, we propose a Sentiment-oriented Transformer-based Variational Autoencoder (So-TVAE) network, which consists of a sentiment-oriented diversity encoder module and a batch attention module, to deal with the above two challenges. 
The proposed sentiment-oriented diversity encoder elegantly combines VAE and random mask mechanism to achieve semantic diversity under sentiment guidance. 
Specifically, we first leverage a Gaussian mixture distribution mapping guided by sentimental information to learn diverse comments with different sentiments. In addition, to effectively avoid VAE ignoring the context information and directly learning the mapping of the latent space, we propose a novel sentiment-oriented random mask encoding mechanism, balancing the learning ability of the model to various modalities, which further improves the performance. 
%

Moreover, a batch attention module is proposed in this paper to alleviate the data imbalance problem existing in live video. We leverage batch-level attention along with multi-modal attention in a parallel way to simultaneously deal with the multi-modal features in batch dimension and spatial dimension. In this way, the proposed batch-level attention module can explore the sample relationships in a mini-batch, introduce the virtual samples to assist the learning of missing sentiment samples, and further improve the quality and diversity of generated comments. 
In addition, to measure the diversity of generated comments, we also propose a novel evaluation protocol for the automatic live video commenting task.
The extensive experiments conducted on two datasets Livebot ~\cite{2018LiveBot}  and VideoIC~\cite{2020VideoIC} ) verify that So-TVAE outperforms the state-of-the-art models by $\textbf{26.4\%}$ in all metrics at average. In order to verify the generalization ability of the proposed model for cross-modal automatic commenting, we further conduct expansion experiments on the image news commenting task, which also achieves state-of-the-art performance.

The main contributions of this work are as follows: 
  \vspace{-0.13cm}  
\begin{itemize}
\item   We propose a Sentiment-oriented Transformer-based Variational Autoencoder (So-TVAE) network, which successfully achieves diverse video commenting with multiple sentiments and multiple semantics. To the best of our knowledge,  we are the first to focus on the diversity of comments in the automatic live video commenting task.

\item We propose a sentiment-oriented diversity encoder module, which elegantly combines VAE and random mask mechanism to achieve semantic diversity and further align semantic features with language and video modalities under sentiment guidance. Meanwhile, we propose a batch-attention module for sample relationship learning, to alleviate the problem of missing sentimental samples caused by the data imbalance.

\item We provide an evaluation protocol for the ALVC  task which can measure both the quality and diversity of generated comments simultaneously, further assisting future research.

\item We perform extensive experiments on two public datasets. The experimental results based on all evaluation metrics prove that our model outperforms the state-of-the-art models in terms of quality and diversity.

\end{itemize}

  \vspace{-0.03cm}  
\section{Related Work}

\subsection{Image Captioning and Video Captioning}
A task similar to live comment generation is caption generation, such as image captioning or video captioning, which aims to convert visual information into natural language descriptions. 
The early work of image captioning mainly adopted template-based methods ~\cite{2010Every,2011Composing,Lalit2014BabyTalk} and retrieval-based methods ~\cite{2014Improving,2015Framing,2015Automatic}.
Then the methods based on deep neural network to generate caption ~\cite{2014Multimodal, 2015Show, wang2018image} became mainstream. ~\cite{xu2015show} proposed an attention mechanism to capture meaningful information in images for the first time.~\cite{2017Knowing,2018Show,cornia2018paying,yuan2020image,yang2020constrained,lu2021chinese} improved and proposed different variants of the attention mechanism. 
Some classical attention mechanisms have been proposed successively, such as ~\cite{2017Bot} proposed bottom-up and top-down attention to selectively focus on spatial image features;   and ~\cite{2019Att} proposed an additional attention module to measure the correlation between attention results and queries.
 Inspired by the success of the Transformer model in machine translation, Transformer-based methods~\cite{8869845,yuan2023adaptive,jiang2021bi,9784827}  are also widely used recently. ~\cite{8869845} was the first to introduce Transformer into image captioning and established a multimodal Transformer model. 
In addition, the methods based on generative adversarial networks ~\cite{2019Compression,2018Latest} and the methods based on reinforcement learning ~\cite{wei2021integrating,2019Self,2017Improved, 2016Self} have been widely used in image captioning.

Video captioning ~\cite{2018Video} aims to generate natural language sentences describing video content. ~\cite{2002Natural,2013Generating} adopted template-based methods to generate captions, which are hard to generate flexible descriptions. 
With the development of deep neural networks, the encoder-decoder framework is widely used to generate descriptions with flexible syntax structure ~\cite{2014Translating,2018Reconstruction,man2022scenario,shi2023learning,dong2023semantic}. 
~\cite{2016Describing} proposed a temporal attention mechanism to model the global temporal structure of videos. ~\cite{2017Learning1} exploited audio features to enrich video representation. 
~\cite{2020Spatio,2020Object,2019Spatio} introduced target detection to generate more accurate descriptions.
Video captioning can also be divided into some subcategories according to specific tasks, such as dense video captioning ~\cite{8807239, 9160989,chen2023retrieval}, sports video captioning ~\cite{8733019,2018Fine}, etc.

There are also some style captioning works ~\cite{2017StyleNet, 2020MSCap, 2020MemCap, mathews2016senticap,chen2018factual}. 
{ 
~\cite{mathews2016senticap} proposed a model consisting of two parallel RNNs to capture and combine the background language and sentimental description. ~\cite{chen2018factual} devised a style-factual LSTM, using two groups of matrices to capture the factual and stylized knowledge and adaptively weighted fusion. These works mostly rely on well-annotated stylized corpora, and use an encoder-decoder framework guided by stylized factors to generate fixed style descriptions. 
Compared with these captioning tasks,  ALVC based on complex and unannotated live videos, aims to generate diverse comments with subjective sentiments, which brings more challenges in data imbalance, multi-modal interaction, and diversity generation. Our model proposes a novel one-to-many mapping framework to parallel capture the multi-modal interacted features in both batch dimension and spatial dimension, to address the challenges.}


  \vspace{-0.23cm}  %
\subsection{Article Commenting, Image Commenting, and Image News Commenting}
Other tasks similar to our work include article commenting and image commenting. For article commenting, ~\cite{2018Automatic} formally proposed the task and contribute a high-quality corpus for further research. 
~\cite{2018Learning} proposed a novel framework combining retrieval and generation methods to generate news comments. ~\cite{2019Automatic} proposed a personalized comment generation network, commenting on social media instead of news articles. ~\cite{li2019coherent} introduced a graph-to-sequence mapping to model the input news as a topic interaction graph.
For image commenting, ~\cite{chen2021nice} formally proposed the task and contributed a large-scale image comment dataset, as well as a pre-training approach to simulate human commenting, which encouraged further work. 
Compared with these single-modal commenting tasks, \textit{i.e.}, article commenting and image commenting, ALVC as a cross-modal commenting task, needs to simulate the interaction between text and video. The dynamic and multimodal information contained in videos and comments presents greater challenges. 

Another similar task in the field of cross-modal commenting is image news commenting, which aims to generate a reasonable and fluent comment with respect to the news image and text.
~\cite{2019Cross} proposed the task for the first time and constructed a related news comments dataset. They also presented a novel co-attention model to capture the internal interaction between multiple modal contents, and explored several representative methods based on the classical neural network and Transformer model.
Different from the simple news scenario in this task, ALVC is directly oriented to real-time interaction based on video content, which is a more complex social media scenario.

  \vspace{-0.23cm}  %
\subsection{Automatic Live Video Commenting}
 Most early works on real-time video comments focused on extracting user preference information from danmaku to help recommendation and tagging, such as keyframe prediction ~\cite{2017Personalized,2019Person}, video tagging ~\cite{2019Stories,2016Reading,2019Time}, recommendation ~\cite{2019Understanding,2019Herding}, and viewer behavior ~\cite{2018Visual,2019Beyond, chen2023weakly}. Different from these tasks, automatic live video commenting aims to generate human-like live comments, which is more suitable to explore the natural language structure and interactivity of comments.
Early work ~\cite{li2016video} proposed a retrieval-based method, which searches and selects the most similar video comment as the response, achieving the related commenting, but weak in semantic understanding.  
 Then autoregressive generation-based work ~\cite{2018LiveBot} was proposed, which devised two basic neural networks: Fusion RNN model and Unified Transformer model to jointly encode the visual and textual information into video comments.
 %
   ~\cite{duan2020multimodal,2020DCA} further explored the interaction between multiple modal information.
 ~\cite{2020VideoIC} proposed a new dataset “VideoIC” based on the consideration of higher comment density and more diverse video types, they also proposed a multimodal and multitask generation model to achieve effective comments generation and temporal relation prediction. ~\cite{2021PLVCG} introduced the post time and video type labels to further consider the temporal interaction between video and comments. ~\cite{2021Knowing} modeled a single unified framework to jointly implement comment generation and highlight detection. 
 However, these works ignore the diversity of comments and only generate a single video comment, resulting in the underutilization of context information.

\vspace{-0.23cm}  
\subsection{One-to-Many Generation}
Some well-known one-to-many generation works are achieved by changing the decoding scheme during inference, such as using a search algorithm as a post-processing step like beam search ~\cite{kumar2004minimum,collobert2019fully}, or reframing generation as a reinforcement learning problem ~\cite{2015Scheduled}. These methods do not make full use of the diversity of target sentences in training, which usually generate similar sentences differed only by a punctuation or a word.
 More directly, there are many ways to model a one-to-many generation from paired data, such as the mixture of experts ~\cite{2019Mixture}, discrete domain encoding ~\cite{2020Target,2017StyleNet} or Variational Autoencoder ~\cite{2014Auto,2014Stochastic,2015Generating}. 
Among these, Variational Autoencoders were proposed by ~\cite{2014Auto} and ~\cite{2014Stochastic}, and were initially applied in the computer vision task. ~\cite{2015Generating}  introduced VAEs into natural language processing for text generation for the first time, then VAEs have been extended by many following works in various language generation tasks ~\cite{2017Latent,2018Unsupervised,2017Learning}. ~\cite{2017Latent} and ~\cite{2018Unsupervised} proposed a discrete latent vector modeling to generate interpretable sentences.  ~\cite{2014Semi, 2017Toward,2017Multi} proposed and developed the semi-VAE for semi-supervised learning.
~\cite{2020A, 2020ControlVAE, 2019Dispersed} made improvements to address the mode collapse problem widely existing in VAEs.
To date, Conditional Variable Autoencoders (CVAEs) have been used in text summarization ~\cite{2019Topic}, image caption ~\cite{2017Diverse}, dialog model ~\cite{2017Learning} and other tasks, our work is the first to apply them to the ALVC task. By addressing the problem of multi-modal learning and sentiment information fusion, our model achieves the generation of diverse comments. 

  \vspace{-0.13cm}  %
\subsection{Data Imbalance Learning}
{
Imbalanced training data brings challenges to model learning, and existing works have explored a lot, by re-weighting ~\cite{cui2019class,cheng2021ba}, data generalization \cite{zhu2018generative}, compositional learning ~\cite{naeem2021learning}, and so on. Among them, ~\cite{cheng2021ba} adopted a batch loss re-weighting strategy to enhance hard-sample learning; ~\cite{naeem2021learning} enhanced the generalization by exploiting the joint compatibility of concepts semantics encoding.
One effective method is to enhance the generalization ability by sample relationship learning ~\cite{lim2022hypergraph,liao2022graph, 2022BatchFormer,hou2019cross}.  ~\cite{lim2022hypergraph} utilized the hyper-graph to model multilateral sample-relation and class-relation. ~\cite{liao2022graph} proposed an intra-neighbor sampling method to construct the batches. ~\cite{2022BatchFormer} first proposed utilizing the Transformer attention mechanism to explore sample relationships through batch interaction. However, these works on sample relationship learning are mostly applied in the image classification task, with only single-modal input. In this work, we proposed a novel batch attention module that effectively combines multi-modal interaction and batch interaction, capturing the sample relationship to alleviate data imbalance for the video commenting task. 
}

\begin{figure*}[!t]
  \vspace{-0.4cm}  
  \setlength{\abovecaptionskip}{0.14cm} 
  \setlength{\belowcaptionskip}{-0.39cm}
\centering
\includegraphics[width=4.5in]{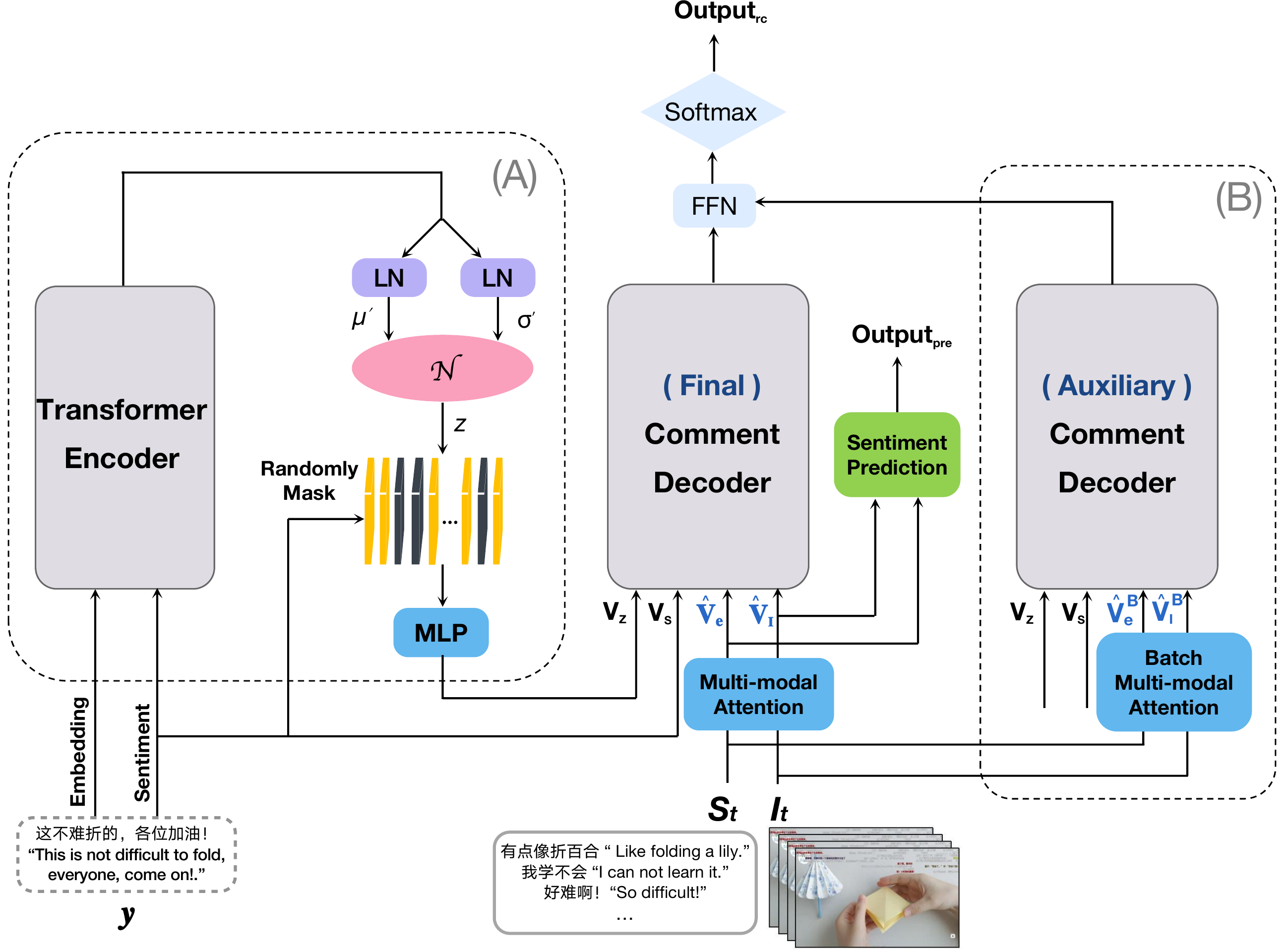}
\caption{Overview of our proposed Sentiment-oriented Transformer-based Variational Autoencoder (So-TVAE) network.
Firstly, the multi-modal encoder and sentiment-oriented diversity encoder (Part A) are used to extract the context features and the diverse sentimental features respectively, then all features are entered into the comment decoder for generation, which is the backbone of So-TVAE. Moreover, a batch attention module (Part B) is used to extract the batch-level context features and paired with the sentimental features to input into a parameter-shared decoder for generation during the training stage. Besides, the sentiment prediction module is used for single comment generation at inference. }
\label{fig_2}
\end{figure*}

\section{Approach}

\subsection{Problem Formulation and Multi-modal Encoder}

Given the video clip $\textbf{V}_t$ with $k$ frames $\textbf{I}_t=\lbrace I_{1},…,I_{t},…,I_{k} \rbrace$ and $m$ surrounding comments $\textbf{S}_t=\lbrace s_{1},s_{2},…,s_{m} \rbrace$ which are firstly concatenated into a word sequence $\textbf{x}_t = \lbrace x_{1},x_{2},…,x_{p} \rbrace$, where $p$ represents the total number of words of $\textbf{S}_t$, automatic live video commenting aims to generate human-like comments at timestamp $t$. 
The overview of the framework is illustrated in Figure \ref{fig_2}.

Specifically, the input frames and surrounding comments are first represented as the initial features with a pre-trained convolution neural network (ResNet) ~\cite{2016Deep} and Long short-term memory network (LSTM) ~\cite{Hochreiter1997Long}, respectively. Then we employ a co-attention module to further enhance feature encoding, and generate the attended visual representations $\hat{\textbf{V}}_\textbf{I}$ and the attended textual representations $\hat{\textbf{V}}_\textbf{e}$:
\begin{align}
V_{i} =\text{ResNet}(I_i), \ \ E_{i} =\text{LSTM}(e(x_{i}),E_{i-1}), \\
 \hat{\textbf{V}}_\textbf{e},\hat{\textbf{V}}_\textbf{I}  = \text{Co-Attention}(  \textbf{V}_\textbf{e}, \textbf{V}_\textbf{I}), \ \  \ \ \ \ \ \ \ \ \ \ \ \ \ \
\end{align}
where $e(\cdot)$ is the word embedding processing. 
Co-attention 
acts on the initial frame features  $\textbf{V}_\textbf{I}=\lbrace V_{1},V_{2},…,V_{k} \rbrace  \in \mathbb{R}^{B\times k \times d}$ and text feature $\textbf{V}_\textbf{e}=\lbrace E_{1},E_{2},…,E_{p} \rbrace \in \mathbb{R}^{B\times p \times d}$ for modeling multi-modal interaction. 

\subsection{Sentiment-oriented Diversity Encoder}

The previous methods of the ALVC task can only achieve one-to-one generation as they encode the source input to a \emph{fixed} feature vector to guide the reconstruction of comments. Therefore, based on the principle of Variational Autoencoder, we propose a diversity encoder that can explicitly model a one-to-many mapping between the latent space and the target comments. By sampling multiple latent vectors $\textbf{z}$ from the trained latent space, our model can generate diverse comments.

\subsubsection{Sentimental Diversity} 

 To make up for the lack of sentiment annotation in the original datasets, a pre-trained sentiment analysis model SKEP ~\citep{2020SKEP} is introduced to evaluate the sentiment of target comment $\textbf{y}$, producing a sentiment label $s$ ($s \in \lbrace0,1,...,N-1\rbrace$,  $N$ is the number of sentiment categories).
 We use one-hot encoding on the sentiment label to obtain sentiment weight vector $\textbf{s}$, followed by an embedding layer to generate the direct sentiment feature representations 
$\textbf{V}_\textbf{s}$:
 \begin{align}
 \textbf{s}&=\text{one-hot}(s),\\
\textbf{V}_\textbf{s} &= \text{Embedding}_\text{s} (\textbf{s}).
\end{align}%

\subsubsection{Semantic Diversity} 
Furthermore, 
we model a one-to-many mapping to learn the diverse semantic information of different sentiments.
In the training stage, the semantic diversity module samples a sentiment-oriented latent vector  $\textbf{z}$ from the latent space with the guidance of target comment $\textbf{y}$.
By making $\textbf{z}$ obey a prior probability distribution, our model can learn the diversity mapping relationship from the latent space to the reconstructed comments. Thus, in the inference stage, our model can generate diverse comments by sampling different sentiment-oriented latent vectors $\textbf{z}$ from the prior distribution.

To effectively integrate the sentimental information and the semantic information, our model explicitly structures the latent space with the guidance of sentimental information.
As the prior distribution of the latent vector $\textbf{z}$ determines how the learned latent space is structured, which is crucial to the model performance, we construct it as a sentiment-based  Gaussian mixture model:
 \begin{align}
p(\textbf{z}\mid\textbf{s}) &=\sum_{j=1}^N s_j \mathcal{N}(\textbf{z}|\mu_j,\sigma^2_j \textbf{I}),\label{2}\\
\textbf{z}_\text{prior} &\sim  \mathcal{N}(\mu, \sigma^{2}; \textbf{s}),
\end{align}%
where $\textbf{I}$ is the identity matrix, 
$\textbf{s}$ is the sentiment weight mentioned before, $\mu_j$ and $\sigma_j$ represent the mean vector and standard deviation of the $j$-th Gaussian distribution. { $\textbf{z}_\text{prior}$ is the prior latent vector, obtained by following the Gaussian mixture sampling of the prior distribution, as defined in Equation \ref{2}.}
 In practice, for all components, we use the same standard deviation $\sigma$.

Concretely, we leverage a Transformer encoder layer to map the target comment $\textbf{y}$ and sentiment weight $\textbf{s}$ to a sampling region of latent space for training. The last hidden state $h_T$ of the Transformer layer is mapped into $N$ mean vectors $\mu_j^{'}$ and $N$ log variances $\log{\sigma^{2}_{j}}^{'}$, using linear network for each:
 \begin{align}
\textbf{H} &=\text{Transformer}({s}_{i},\textbf{s},\textbf{y}),\\
\mu_j^{'},\log{\sigma^{2}_{j}}^{'} &=\text{LN}_\text{j}(h_T) , \quad   \text{for}\  j=1,...,N \\
\textbf{z}_\text{post} &\sim  \mathcal{N}(\mu^{'}, \sigma^{2'}; \textbf{s}),
\end{align}%
where $\textbf{z}_\text{post}$ is the posterior latent vector, obtained by following the Gaussian mixture sampling mode of prior distribution in Equation \ref{2}. {We adopt the reparameterization trick ~\cite{2017Diverse, 2019Dispersed} to ensure gradient continuity in the Gaussian mixture sampling, as detailed in Appendix \ref{repar}.}

\subsubsection{Sentiment-oriented Random Mask} 
Contrary to the posterior collapse problem of VAE existing in other tasks, our model confronts a new \emph{mode imbalance} problem where the model tends to rely heavily on the latent vector rather than the source input. 
Different from other tasks, the sentiment-guided ALVC is more difficult to reconstruct the target comments from the source input. As the latent space is infinite, the model tends to ignore the input information and directly learns the mapping between the encoding region of latent space and the reconstructed comments, which is useful in training, but meaningless at inference.Therefore, we propose a masked encoding mechanism to balance the learning ability of the model to various features by sentimental masking the posterior latent vector $\textbf{z}_\text{post}$ on the random mask region $m_\lambda$. With a linear network to encode the latent vector, the multi-semantic sentiment feature representation is calculated as:
 \begin{align}
\textbf{V}_\textbf{z}=\text{Encoding}_\text{z}((1-m_\lambda)\odot \textbf{z}_\text{post} + m_\lambda \odot \textbf{z}_\text{prior}),
\end{align}%
where $m_\lambda$ is a binary mask region, which regional proportion is determined by the mask ratio $\lambda$, $\odot$ is element-wise multiplication. 
$\textbf{z}_\text{prior}$ and $\textbf{z}_\text{post}$ correspond to the prior and posterior hidden vectors obtained above.

\subsection{Batch attention Module}
To address the problem of sentimental sample missing caused by \emph{data imbalance}, inspired by the batch interaction thought of~\cite{2022BatchFormer}, we devise a novel batch attention module to explore the sample relationship to introduce virtual samples assisting the learning of missing sentiments.
{The proposed batch attention module makes a targeted improvement toward to the input characteristics of video commenting task, both on algorithm structure and attention mechanism, to achieve a comprehensive utilization of the input information with multi-modal and multi-sample.  By parallelly leveraging the batch multi-modal attention and the original multi-modal attention based on textual-visual co-attention, the model achieves the processing of multi-modal features in both batch dimension and spatial dimension.} 
In this manner, the proposed module can mitigate the gap between samples and model their relations.

Given a batch sample $\textbf{X}=(X_1, X_2,..., X_B)$  with size $B$, the batch attention module implicitly augments $B-1$ virtual samples for each sample $X_{i}$ by modeling the relationship among samples in the mini-batch, resulting in an improvement on data scarcity. Specifically, we adopt a co-attention module acting on the input features to obtain the batch-attended textual and visual features:
 \begin{align}
\hat{\textbf{V}}^\textbf{B}_\textbf{e} , \hat{\textbf{V}}^\textbf{B}_\textbf{I} = ( \text{Co-Attention}_\text{B} (\textbf{V}^\textbf{B}_\textbf{e} , \textbf{V}^\textbf{B}_\textbf{I}) )^\text{T},
\end{align}
where $ \textbf{V}^\textbf{B}_\textbf{e}  \in \mathbb{R}^{p\times B \times d}$ and $\textbf{V}^\textbf{B}_\textbf{I} \in \mathbb{R}^{k\times B \times d}$ are reshaped from the original frame feature and text feature, to enable the co-attention module working on the batch dimension. {Then the batch-attended features are inputted into the auxiliary decoder for decoding, playing the same role as the multi-modal attended features which are inputted into the final decoder. In the training stage, the batch attention module directly encodes and utilizes the sample relationship to enhance decoding.}

In the inference stage, considering the single-sample input, the batch attention module cannot be used directly. {Directly removing the batch attention will result in the loss of the learned sample relationship. Therefore, we propose a simple but effective method, by sharing the parameters/weights between the auxiliary decoder and the final decoder, which not only helps to partly keep the consistency of the decoding features with and without batch attention, but also transfers the sample relationship learned by batch attention into the model backbone. In this case, the model can always benefit from the learned sample relationship, whether with or without batch attention.}

\subsection{Decoder}
\subsubsection{Sentiment-Prediction} 

To predict the sentiment most expected by the viewers for a video clip, we design a sentiment-prediction module to predict the most likely comment sentimental types based on the visual and textual context. The predicted sentiment label $\hat{s}$ is calculated as:
 \begin{align}
\textbf{V}_\textbf{pre} = \text{LayerNorm}( \textbf{W}_\textbf{I} \hat{\textbf{V}}_\textbf{I} +\textbf{W}_\textbf{e}\hat{\textbf{V}}_\textbf{e} ) ,\\
p(\hat{s} \mid \textbf{I}_t ,\textbf{x}_t) = \text{Sigmoid} (\textbf{W}_\textbf{pre}\textbf{V}_\textbf{pre}),
\end{align}
where $\textbf{W}_\textbf{I}, \textbf{W}_\textbf{e} \in \mathbb{R}^{d\times d_{pre}}$ and $\textbf{W}_\textbf{pre} \in \mathbb{R}^{d_{pre}\times N}$ are learnable weight matrixes. 
To compare with the previous single-generation works, the sentiment label $\hat{s}$ predicted based on context can be used for generation at inference, replacing the sentiment label $s$ from target comments.

 \subsubsection{Comment Decoder} 
The goal of the comment decoder is to generate comments for the target timestamp based on context information and diverse semantic and sentimental information. More specifically,  we decode the word $\hat {{y}}_{t} $ at every timestep $t$ with a Transformer decoder layer followed by a softmax layer: 
\begin{align}
w_t =\text{Transformer} ({y}_{t}, \textbf{y}_\textbf{\textless{t}},\hat{\textbf{V}}_\textbf{I} , \hat{\textbf{V}}_\textbf{e},  \textbf{V}_\textbf{s} ,\textbf{V}_\textbf{z}) ,\\
p(\hat {{y}}_{t} \mid y_0,…,y_{t-1}, \textbf{I}_t ,\textbf{x}_t ,s,\textbf{z}) =\text{ Softmax}(\textbf{W}w_t) ,
\end{align}%
where $y_t$  indicates the $t$-th token in comments $\textbf{y}$ and $\textbf{y}_\textbf{ \textless t}=\lbrace y_0,…,y_{t-1}\rbrace$ denotes the masked comment.   
Inside the comment decoder, there are five multi-head attention blocks, using $y_t$ as the query to attend to $\textbf{y}_\textbf{ \textless t},\hat{\textbf{V}}_\textbf{I} , \hat{\textbf{V}}_\textbf{e} , \textbf{V}_\textbf{s} $ and $ \textbf{V}_\textbf{z}$, respectively.
The auxiliary comment decoder has the same structure, but using $y_t$ as the query to attend to $\textbf{y}_\textbf{ \textless t}, \hat{\textbf{V}}^\textbf{B}_\textbf{I} , \hat{\textbf{V}}^\textbf{B}_\textbf{e} , \textbf{V}_\textbf{s} $ and $ \textbf{V}_\textbf{z}$, respectively.

\subsection{Loss function}
The traditional encoder-decoder model is realized by maximizing the log-likelihood function of generated comment $\hat{\textbf{y}}$. As our generation model is controlled by latent vector $\textbf{z}$, the objective is  changed to maximize:
 \begin{align}
p( \hat{\textbf{y}}) = \int	_\textbf{z} p( \hat{\textbf{y}}  \mid \textbf{z} ) p( \textbf{z} ) d\textbf{z}.
\end{align}%
As it is impossible to traverse all latent vectors $\textbf{z}$ for integration, by referring to the mathematical derivation in the Variational Autoencoder~\citep{2014Auto}, we use the Evidence Lower Bound (ELBO) to approximate the log-likelihood function of generated comment $ \hat{\textbf{y}} $:
 \begin{align}
\log{p( \hat{\textbf{y}})} \geq E_{q(\textbf{z}\mid \hat{\textbf{y}})} [\log{ p(\hat{\textbf{y}} \mid \textbf{z} )} ] -D_{KL}[q(\textbf{z}\mid \hat{\textbf{y}} ),p(\textbf{z})],
\end{align}%
where $q(\textbf{z}\mid \hat{\textbf{y}})$ corresponds to the probability distribution obtained by the diversity encoder and is used to approximate the posterior probability distribution $p(\textbf{z}\mid \hat{\textbf{y}} )$ of the decoder.

Since the model is also affected by the video frames $\textbf{I}_t$, surrounding comments $\textbf{x}_t$, and generated or selected sentiment type $s$, the objective function is changed to:
\begin{align}
\log{p( \hat{\textbf{y}}\mid \textbf{I}_t ,\textbf{x}_t,s)} \geq E_{q(\textbf{z}\mid  \hat{\textbf{y}},\textbf{I}_t ,\textbf{x}_t,s)}[\log{ p(\hat{\textbf{y}} \mid \textbf{z} ,\textbf{I}_t ,\textbf{x}_t,s}) ] - \nonumber\\
D_{KL}[q(\textbf{z}\mid \hat{\textbf{y}},\textbf{I}_t ,\textbf{x}_t, s ),p(\textbf{z}\mid \textbf{I}_t ,\textbf{x}_t,s)].
\end{align}%
The first item encourages the model to generate higher quality comments, which is the loss of model reconstruction, named as $loss_{rc}$. The second item encourages the latent vector distribution obtained by training to be as close as possible to the prior distribution $p(\textbf{z}\mid \textbf{I}_t ,\textbf{x}_t,s)$, i.e., $p(\textbf{z}\mid s)$, which is the KL distance between them: 
\begin{align}
    loss_{z}&=D_{KL}[q(\textbf{z}\mid\hat{\textbf{y}},\textbf{I}_t ,\textbf{x}_t,s   ),p(\textbf{z}\mid s)] \nonumber\\
    &=\sum_{j=1}^N \log({\frac{\sigma_j}{\sigma_{j}^{'}}})+\frac{{\sigma_j^{'}}^2+{\mid\mid \mu_{j}^{'}-\mu_j \mid\mid}_2^2}{2\sigma_{j}^2} -\frac{1}{2}. 
\end{align}%
Moreover, considering the  sentiment-prediction module, we adopt cross-entropy loss to optimize:
\begin{align}
 loss_{pre} = -\log{p_{\theta} (\hat{s} \mid s)},       
\end{align}%
where $\theta$ corresponds to model parameters, $\hat{s}$ and $s$ are the predicted sentiment and ground truth sentiment respectively. 

In short, the total loss function of the model is defined as: 
\begin{align}
\mathcal{L}= loss_{rc} +\beta \cdot loss_{z} + \gamma \cdot loss_{pre},
\end{align}%
where $\beta$ and $\gamma$ are hyper-parameters.

\section{Experiment}

\subsection{Experiment Setting}
The experiments are performed on two live video commenting datasets Livebot ~\cite{2018LiveBot} and VideoIC ~\cite{2020VideoIC} which are all collected from a popular online video website, \text{BiliBili}\footnote{\url{https://www.bilibili.com}. More details for the dataset are provided in Appendix \ref{data}}. 
In all experiments, we consider the Transformer layers with $6$ blocks, $8$ attention heads, and set the hidden size of the multi-head attention layers and the feed-forward layers to $512$ and $2,048$. 
Our model is optimized by Adam optimizer~\citep{2014Adam} with the base learning rate set to $1e-4$ and $3e-5$ on datasets Livebot and VideoIC respectively, and  decayed by $1/4$ every $2$ epochs after $4$ epochs. 
The batch size is $128$ and dropout rate is $0.1$. 
The number of selected video frames and surrounding comments $k$ and $m$ are both set to $5$.
For the joint loss function,  the corresponding parameters are set as: $\beta=2$ and $\gamma=0.3$. We adopt $3$ as the number of layers of the co-attention module.
The standard deviation of each prior distribution is fixed at $0.2$. Unless otherwise stated, the sentiment category $N$ is set to $3$, as $N = 3$ exactly corresponds to the three sentiments of negative, neutral, and positive.
 
\subsection{Evaluation Metrics.} On the one hand,  following the previous work, we use the retrieval-based evaluation metrics proposed by ~\cite{2018LiveBot} for the quality evaluation of single comment generation. Since a time stamp contains multiple various comments, which makes the reference-based metrics not applicable, the evaluation is formulated as a ranking problem.
The model is asked to rank a set of candidate comments based on the log-likelihood score.
We provide a candidate comment set that consists of diverse ground-truth comments and three types of improper comments as follows: 
\begin{itemize}
\item 
 $\textbf{Correct}$: the ground-truth comments of the corresponding video and corresponding moment provided by human.
\item 
 $\textbf{Plausible}$: $50$ comments most similar to the video title which are measured based on the cosine similarity of their TF-IDF values.
\item
 $\textbf{Popular}$: $20$ most popular comments from the dataset which carry less information, such as ``Great”, ``hahaha”.
\item
 $\textbf{Random}$: after selecting the correct, plausible, and popular comments, randomly select some comments to make sure the number of the candidate set contains 100 comments.
\end{itemize}

The model ranks the ground-truth comments higher with a higher generation probability, reflecting better model performance. We measure the ranking results with the following metrics: 
 \begin{itemize}
\item$\textbf{Recall@k}$: the proportion of ground-truth comments in the top-k sorted comments, higher is better.

\item$\textbf{Mean Rank}$: the mean rank of the ground-truth comments, lower is better.

\item$\textbf{Mean Reciprocal Rank}$: the mean reciprocal rank of the ground-truth comments, higher is better.

\end{itemize}

On the other hand, inspired by some one-to-many generation works ~\cite{2020Target,zhu2018texygen}, we propose an evaluation protocol for ALVC to measure both the quality and diversity of our generations. For the $i$-th sample which contains multiple ground-truth comments $[ r_i^{1},r_i^{2},…,r_i^{K} ]$, our model can generate corresponding $N$ comments $[ h_i^{1},h_i^{2},…,h_i^{N}]$. 
To measure the quality and diversity of our generations, we define:
\begin{align}
\textbf{BLEU}_\textbf{ref} &=\text{BLEU}(\lbrace h_i^{j}, [r_i^{1}, …, r_i^{K}] \rbrace, \ _{ 1\leq j \leq N, \  1\leq i \leq P} ),\\
\textbf{BLEU}_\textbf{self}&=\text{BLEU}(\lbrace h_i^{j}, h_i^{k} \rbrace, \ _{1\leq k\leq N \ and \   k \neq j, \ 1\leq i \leq P }),
\end{align}%
where $P$ is the total number of test samples. 

As $\text{BLEU}_\text{ref}$  and $\text{BLEU}_\text{self}$ compute the correlation between generations and references, and the correlation between generations respectively, a model that has a low $\text{BLEU}_\text{self}$  score with a high $\text{BLEU}_\text{ref}$  score performs well in generative diversity and quality. Different from the previous quality metrics, $\text{BLEU}_\text{ref}$  measures the average quality of multiple generated comments, and $\text{BLEU}_\text{self}$  corresponds to the generative diversity. For comparison, the previous one-to-one generation models are paired with beam search to generate multiple comments with beam size set to $3,5$ and $10$, displaying the best results.

\subsection{Comparison with State-of-the-Art}
In order to verify the effectiveness of the proposed model, we conduct experiments to compare our So-TVAE with the state-of-the-art methods on two datasets. The strong baselines are as follows:
 \begin{itemize}
\item$\textbf{S2S}$: the model includes two encoders and one decoder, which is similar to ~\cite{2017Attention}. Two encoders encode the video frames and surrounding comments respectively, and the decoder generates comments based on the concatenated outputs of the encoders. 

\item$\textbf{Fusional RNN}$ \textbf{(F-RNN)} ~\cite{2018LiveBot}: the model uses LSTMs and attention mechanism to construct encoder and decoder.

\item$\textbf{Unified Transformer}$ \textbf{(U-Trans)}  ~\cite{2018LiveBot}: the model employs multiple Transformer modules to construct encoder and decoder.

\item$\textbf{MML-CG}$  ~\cite{2020VideoIC}: the model is based on multimodal multitasking learning to capture the temporal relation and interaction between multiple modalities for comments generation. 

\item$\textbf{Matching Transformer}$ \textbf{(M-Trans)} ~\cite{duan2020multimodal}: the model constructs a multimodal matching network based on the Transformer framework to capture the relationship among text, vision, and audio.
\end{itemize}

For our sentiment-oriented model, to further explore the impact of sentiment, we conduct experiments with different sentiment categories ($3$, $5$, and $10$).
 Tabel \ref{tabel1} shows the comparison results of the baseline models and our proposed model. 
Our proposed \text{So-TVAE} exhibits better performances on both two datasets, showing an average improvement of $\textbf{26.4\%}$ in all metrics, which verifies the superiority of diversity learning on live video commenting. 
It is worth noting that our sentiment-oriented model does not simply improve with the increase of sentiment categories. The model achieves the best performances on Livebot dataset with $N=5$  and VideoIC dataset with $N=10$, meaning that the best effects will be obtained when the fine granularity of sentiment is matching the sentimental distribution of the dataset.

\begin{table*}[]

  \setlength{\abovecaptionskip}{0.2cm} 
    \vspace{-0.27cm}  
\centering
\caption{Performance comparisons with the state-of-the-art methods, where R@k, MR, MRR, Br@k, Bs@k are short for Recall@k, Mean Rank, Mean Reciprocal Rank, {BLEU}$_{ref}$ and {BLEU}$_{self}$. R@k, Br@k and Bs@k are reported as percentage (\%). $^\downarrow$ indicates that the performance is better with a lower score. $_{3, 5, 10}$ annotates the results when setting $N=3, 5,10$. }
\label{tabel1}
\renewcommand\arraystretch{1.1}
\resizebox{1\linewidth}{!}{
\begin{tabular}{@{}l|lllllll|lllllll@{}}
\toprule                                                                                                                                             
 & \multicolumn{7}{c}{\small{Livebot}}                          & \multicolumn{7}{c}{\small{VideoIC}}               \\ 
\small{Model}   & \small{R@1} & \small{R@5} &\small{R@10} & \small{MR}$^\downarrow$ & \small{MRR} & \small{Br@1} & \small{Bs@1}$^\downarrow$ & \small{R@1}& \small{R@5} & \small{R@10} & \small{MR}$^\downarrow$ & \small{MRR} & \small{Br@1} & \small{Bs@1}$^\downarrow$ \\ \midrule
S2S ~\cite{2017Attention}    & 10.70 & 25.53 & 32.14 & 20.33 & 0.196 & 9.705 & 88.90 &25.65&54.08&63.85&12.80&0.384&7.056&84.23      \\
F-RNN  ~\cite{2018LiveBot}   & 17.25 & 3796  & 56.10 & 16.14 & 0.271 & 3.549 & 94.72 &25.38&51.02 &64.36&11.73&0.400&13.09&93.81      \\
U-Trans~\cite{2018LiveBot}   & 18.01 & 38.12 & 55.78 & 16.01 & 0.275 & 6.760 & 92.19 &26.34 & 54.66 &64.37 &12.66 &0.390&15.07&91.43     \\
MML-CG ~\cite{2020VideoIC}  & 18.57 & 43.80 & 54.09 & 15.37 & 0.312 &9.027 & 91.21&  27.50  & {56.12}   & 65.68 &12.21  &0.402 &17.73 & 91.17    \\
M-Trans ~\cite{duan2020multimodal}& 22.71 & 46.71 & 62.87 & 11.19 & 0.352 & --    & --    & --  & --  & --   & -- & --  & --   & --   \\  \hline
So-TVAE$_{3}$ &{25.88} & {50.64} & {65.68} & {11.10} &{0.384} & {15.13} &{31.33}             &  29.13 & 54.91 & 69.37 & 9.712 & 0.419  &\textbf{23.15} &18.90 \\
So-TVAE$_{5}$ &\textbf{27.07}&\textbf{52.97}&\textbf{67.44}&\textbf{10.54}&\textbf{0.397}&\textbf{16.26}&29.24      &27.50 & 54.73 & 69.93 & 9.632 & 0.407 &20.61 &14.30\\
So-TVAE$_{10}$&24.04&51.46&66.83&10.73&0.374&14.57&\textbf{27.45 }          &\textbf{29.59} & \textbf{56.39} & \textbf{70.34} & \textbf{9.431} & \textbf{0.425} &19.87 &\textbf{11.88} \\\bottomrule 
\end{tabular}}
  \vspace{-0.22cm}  
\end{table*}

\subsection{Human Evaluation}
\begin{figure}
  \setlength{\abovecaptionskip}{0.2cm} 
      \vspace{-0.27cm}  
\begin{minipage}{.45\linewidth}
\captionof{table}{Human evaluation result on the Livebot test split.}
\label{human}
\centering
\resizebox{0.8\linewidth}{!}{
\begin{tabular}{lllll}
\toprule  
\bf{Model}   & \bf{Flu.} & \bf{Rel.} & \bf{Cor.}  &\bf{Div.} \\  \midrule
F-RNN  &2.44 &2.52 &2.40 &1.32\\
U-Trans&3.44 &3.28 &3.52 &1.48\\
MML-CG  &3.82 &3.44 &3.84 &1.42\\
So-TVAE     &\textbf{4.06} &\textbf{3.72} &\textbf{4.32} &\textbf{3.88} \\  \hline
Human    &4.64 &4.12 &4.52 &4.24\\\bottomrule
\end{tabular}}
\end{minipage}
\hfill
\begin{minipage}{.48\linewidth}
\centering
\includegraphics[width=5cm]{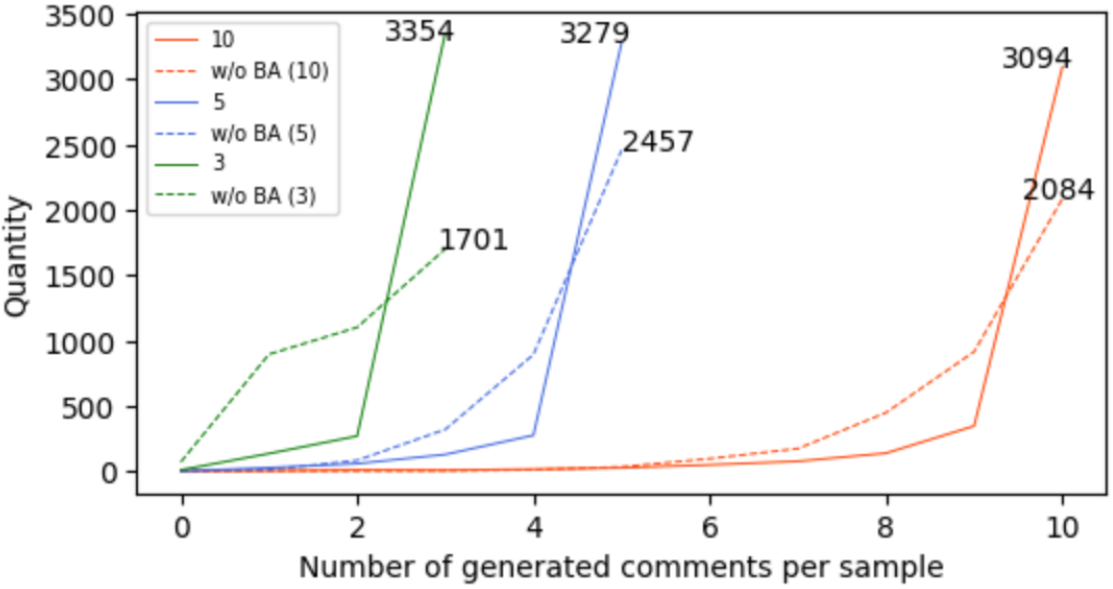}
\caption{Quantity distribution curve of meaningful generated comments.}
\label{batch}
\end{minipage}%
    \vspace{-0.44cm}  
\end{figure}

Human evaluation helps to test the ability of models to generate human-like comments. The generated comments are evaluated from the following four aspects:
 \begin{itemize}
\item{\textbf{Fluency}}: the fluency of comments.
\item{\textbf{Relevance}}: the relevance of comments to video content.
\item{\textbf{Correctness}}: the confidence that the comments are made by humans.
\item{\textbf{Diversity}}: the diversity of multiple generated comments.
\end{itemize}
We randomly select $200$ video clips, which include $1,000$ comments from three baseline models, our So-TVAE, and ground-truth. Three human annotators are asked to score the selected instances on four metrics in the range of $[1, 5]$, with the higher the better. Finally, we take the average score of all annotators as the human evaluation results.
As can be seen from Table \ref{human}, our proposed So-TVAE achieves better performance in all aspects, especially in correctness and diversity. The great improvement in correctness and diversity indicates that our model can generate more human-style comments with diversity learning.

\subsection{Ablations Study}
To fully examine the contribution of each design in our model, we conduct ablation studies to compare different variants of So-TVAE. 
The variants of each module are shown below:
 \begin{itemize}
\item \textbf{Diversity Encoding}: 1) the base model without any diversity components; 2) \text{SMD}: the base model with \textbf{S}e\textbf{M}antic \textbf{D}iversity, which 
uses the standard normal distribution as the prior distribution without sentimental distinction; 3) \text{SEND}: 
 the base model with \textbf{SEN}timental \textbf{D}iversity, which simply encodes the sentimental information as the input to the decoder. 
  \item \textbf{Mask}:  the models with different mask radio $\lambda$ ($\lambda=\%30$ for Full model), where $\lambda=0$ means without masked encoding. 
  
 \item \textbf{Batch Attention Module}:   1) the base model without any batch attention module; 2) the model which uses the original BatchFormer module proposed in ~\cite{2022BatchFormer} to model the sample relationship; 3) the models which use the batch attention module with different batch size (Batch-size=$128$ for Full model).
 \item \textbf{Co-attention}: the model without co-attention. More comparative details are in Appendix \ref{coa1}.
{
 \item \textbf{Loss function} :  the model with different hyper-parameters $\beta$ and $\gamma$ in the loss function ($\beta=2, \gamma=0.3$ for Full model). More comparative details are in Appendix \ref{loss}. }
  \end{itemize} 
According to the results shown in Table \ref{Ablation}, we have the following observations.


\begin{table*}[]
  \setlength{\abovecaptionskip}{0.2cm} 
    \vspace{-0.27cm}  
\caption{Ablation Study on datasets Livebot and VideoIC, where R@k, MR, MRR, Br@k, Bs@k are short for Recall@k, Mean Rank, Mean Reciprocal Rank, {BLEU}$_{ref}$ and {BLEU}$_{self}$. R@k, Br@k and Bs@k are reported as percentage (\%). $^\downarrow$ indicates that the performance is better with a lower score.}
\label{Ablation}
\centering
\renewcommand\arraystretch{1.1}
\resizebox{0.85\linewidth}{!}{
\begin{tabular}{@{}l|l|l|lllllll@{}}
\toprule
   
\small{Dataset  }                 &  \small{Module}               & \small{Variant }    & \small{R@1} & \small{R@5} & \small{R@10} & \small{MR}$^\downarrow$ & \small{MRR} & \small{Br@1} & \small{Bs@1}$^\downarrow$ \\ \midrule
\multirow{14}{*}{Livebot} & \multirow{3}{*}{Diversity}   & w/o        &24.34&47.21&62.47  &12.36&0.363& 8.166&93.04     \\
                          &                              & SEND       &24.39 &47.24 &62.83 &12.61 & 0.362 &14.85 &100.0     \\
                          &                              & SMD       &25.22 &50.50 &64.76 &11.65 &0.382 &15.97 &89.39     \\ \cline{2-10} 
                          & \multirow{3}{*}{Mask}        & $\lambda$=0\%       &24.50 &49.18 &63.96 &11.97 &0.370 & 14.88&\textbf{28.53}      \\
                          &                              & $\lambda$=15\%     &25.21 &50.42 &64.76 &11.48 &0.379& 15.02 & 30.48     \\
                          &                              & $\lambda$=45\%     & 24.84 &50.61 &65.60 &11.28 &0.378 & \textbf{16.78}&35.16       \\\cline{2-10} 
                          & \multirow{4}{*}{Batch-attention} & w/o         &21.39 & 45.12 &59.66&12.83 &0.336&12.66&30.51      \\
                          &                              & BatchFormer  & 22.93 &47.64 &61.94 &12.33 &0.354& 12.66 & 31.55      \\
                          &                              & Batch-size =64     & 24.84 & 50.58 & 64.30 & 11.53 & 0.376 &16.59 &34.20     \\ 
                          &                              & Batch-size =256     &\textbf{26.04} & \textbf{51.99} & \textbf{67.04} & \textbf{10.67} & \textbf{0.389} &14.05 &32.67      \\  \cline{2-10} 
                          & \multirow{1}{*}{Co-attention}      & w/o        &8.678 &20.04 &34.34 &22.12 &0.171&7.321&--      \\  \cline{2-10} 
                          & \multirow{8}{*}{Loss function} &$\beta$=0.5, $\gamma$=1.0  &24.74&	49.30&	65.04&	11.30&	0.380&	13.87&	32.95\\
                          &                                               & $\beta$=1.0, $\gamma$=1.0 &25.06&	50.16&	65.17&	11.18&	0.381&	14.05&	32.63 \\
                          &                                               & $\beta$=1.5, $\gamma$=1.0 &25.42&	50.23&	65.23&	11.15&	0.382	&14.73&	31.48 \\
                          &                                               & $\beta$=2.0, $\gamma$=1.0 &25.59&	50.28&	65.36&	11.16&	0.382&	14.82&	31.40 \\
                          &                                               & $\beta$=2.5, $\gamma$=1.0 &25.29&	50.08&	65.22&	11.13&	0.381&	14.68&	31.61\\         
                          &                                               & $\beta$=2.0, $\gamma$=0.1 &25.23&	50.56&	65.03&	11.11&	0.382&	14.79&	32.24\\                  
                          &                                               & $\beta$=2.0, $\gamma$=0.5 &25.74&	50.53&	65.39&	11.13&	0.383&	15.03&	31.42\\        
                          &                                               & $\beta$=2.0, $\gamma$=1.5 &24.81&	49.71&	64.23&	11.50&	0.367&	14.35&	32.51\\        \cmidrule(l){2-10} 
                          & Full model                   & So-TVAE   &{25.88} & {50.64} & {65.68} & {11.10} &{0.384} & {15.13} &{31.33}          \\ \hline \hline
\multirow{14}{*}{VideoIC} & \multirow{3}{*}{Diversity}   & w/o       &28.44 &52.10 &66.39 &10.73 &0.399 &10.79&96.51    \\  
                          &                              & SEND     &28.73 & 53.48 & 67.70 & 10.30 & 0.412 &23.54 &100.0   \\
                          &                              & SMD     &29.20 & 54.47 &68.41 & 10.10 & 0.418 &24.73 &87.14   \\  \cline{2-10} 
                          & \multirow{3}{*}{Mask}        & $\lambda$=0\%     &28.48 & 54.44 & 67.87 & 9.964 & 0.403   &23.07 &\textbf{17.58} \\
                          &                              & $\lambda$=15\%    &29.22 & 55.27 & 68.89 & 9.798 & 0.421&  23.48&18.18  \\
                          &                              & $\lambda$=45\%     & 29.05 & 55.08 & 69.40 & 9.614 & 0.419 & \textbf{25.15} &18.88 \\  \cline{2-10} 
                          & \multirow{2}{*}{Batch-attention} & w/o        &21.39 & 45.12 &59.66&12.83 &0.336&12.66&30.51     \\
                          &                              & BatchFormer &29.00& 53.42 & 67.43 & 10.55 & 0.412 &22.75 &20.35 \\ 
                          &                              & Batch-size =64     & 28.06 & 53.26 & 68.86& 9.900 & 0.410 &24.58 &24.74     \\ 
                          &                              & Batch-size =256    &\textbf{29.58} & \textbf{55.35} & \textbf{69.67} & \textbf{9.538} &\textbf{0.423}&{23.66}&18.79    \\  \cline{2-10} 
                          & \multirow{1}{*}{Co-attention}      & w/o       &16.98&38.08&53.26&15.69&0.284&14.38& --   \\  \cline{2-10} 
                          & \multirow{8}{*}{Loss function} &$\beta$=0.5, $\gamma$=1.0   &28.02&	53.04&	68.02&	10.32&	0.384&	21.83&	20.31\\
                          &                                               & $\beta$=1.0, $\gamma$=1.0  &28.24&	53.31&	68.17&	10.18&	0.386&	22.08&	19.96\\
                          &                                               & $\beta$=1.5, $\gamma$=1.0 &28.33	&53.76&	68.34&	10.13&	0.393&	22.21&	19.75\\
                          &                                               & $\beta$=2.0, $\gamma$=1.0 &28.46	&53.84&	68.46&	10.01&	0.398&	22.39&	19.52\\      
                          &                                               & $\beta$=2.5, $\gamma$=1.0  &28.42	&53.62&	68.12&	10.04&	0.395&	22.17&	19.81\\                  
                          &                                               & $\beta$=2.0, $\gamma$=0.1 &28.92	&54.77&	69.03&	9.815&	0.401&	22.86&	19.23\\        
                          &                                               & $\beta$=2.0, $\gamma$=0.5 &28.74	&54.42&	68.88&	9.974&	0.410&	22.75&	19.16\\        
                          &                                               & $\beta$=2.0, $\gamma$=1.5 &27.92	&53.65&	68.05&	10.12&	0.392&	21.87&	19.79\\        \cmidrule(l){2-10} 
                          & Full model                   & So-TVAE        &  29.13 & 54.91 & 69.37 & 9.712 & 0.419  &{23.15} &18.90    \\ \bottomrule
\end{tabular}}
    \vspace{-0.27cm}  
\end{table*}

\subsubsection{Contribution of Diversity Encoding}
Obviously, the semantic diversity module and the sentimental diversity module already respectively bring an improvement with respect to the base model. 
It is worth noting that it is not feasible to directly use sentimental information to improve generation diversity (SEND). The model tends to ignore sentimental information and degenerate into a one-to-one generation model without sentimental distinction, resulting in exactly the same comments of different sentiments ($\text{Bs@1}=100.0$).  
By fusing the sentimental information with our semantic diversity module, our model can achieve better performance with more fine-grained comment generation. This also proves the rationality of our sentiment-oriented Gaussian mixture distribution modeling for the latent space, which makes full use of sentimental diversity.

\subsubsection{Contribution of Random Mask Module}
From Table \ref{Ablation}, we can observe that the model shows better generation diversity (\text{Bs@k}) with a lower mask ratio $\lambda$, and shows better average quality (\text{Br@k})  with a higher mask ratio $\lambda$ for multiple generated comments, which is consistent with our conjecture that the mask encoding module can effectively balance the learning ability of the model for source input feature and diversity feature. 
By setting the appropriate mask ratio to capture a balance between diversity and average quality, the model achieves the best overall performance with $\lambda = 30\%$ (Full model).

\subsubsection{Contribution of Batch Attention Module}

The models with batch interaction substantially outperform the model without it, indicating the effectiveness of sample relationship learning for reconstruction. 
{Compared with the original BatchFormer module, the batch attention shows an average improvement of $9.12\%$ in all metrics, which proves the superiority of the proposed batch attention in video commenting tasks. }
In addition, Table \ref{Ablation} also shows a tendency that the batch attention module exhibits better performance with the larger batch size (Batch-size=$256$).

Furthermore, to further demonstrate the effectiveness of the proposed batch-attention module on data imbalance problems, we carry out extensive experiments. For sentiment-guided live video commenting, an unfavorable situation of data imbalance is that with the absence of comment samples of corresponding sentiment, the model is difficult to learn this type of comment information and tends to generate an empty sentence. Thus, we use the model with and without batch attention module to generate $N$ ($3$, $5$, and $10$) comments for all $3,768$ samples in the Livebot test split, exploring the distribution of meaningful (not empty) comment numbers. 

Figure \ref{batch} shows the distribution results. Obviously, we can observe that the model with batch attention module can generate more meaningful comments than without it. Taking the 10 sentiments situation as an example, $3,094$ samples can generate comments for each sentiment using the full model, while only $2,084$ samples can generate all sentimental comments without batch attention. 
These results further prove that our proposed batch-attention module can effectively alleviate the problem of data imbalance in sentiment-guided live video commenting.

\subsubsection{Contribution of Co-attention Module }
Compared with the model without co-attention, the full model shows a huge improvement in performance, validating the effectiveness of the co-attention module for multi-modal interaction.

\subsubsection{Contribution of hyper-parameters in loss function }
{
By implementing the control variable method on hyper-parameters $\beta$ and $\gamma$ in sequence, we obtain a suitable setting for hyper-parameters of $\beta=2$ and $\gamma=0.3$. This conclusion holds for both datasets. In addition, within a wide range of hyper-parameters ( $\beta$ in $[1, 2]$, $\gamma$ in $[0.3, 1]$), the model exhibits roughly the same performance, indicating the strong robustness of our model on hyper-parameters.
}



\begin{table*}[]
  \setlength{\abovecaptionskip}{0.2cm} 
    \vspace{-0.27cm}  
\centering
\caption{Comparison of comment generative diversity, where Br@k and Bs@k are short for {BLEU}$_{ref}$ and {BLEU}$_{self}$. $_{3, 5, 10}$ annotates the results when setting $N=3, 5,10$. All values are reported as percentage (\%). $^\downarrow$ indicates that the performance is better with a lower score.}
\label{diversity}
\resizebox{0.8\linewidth}{!}{
\begin{tabular}{@{}l|llll|llll@{}}
\toprule                                                                                                                                     
 & \multicolumn{4}{c}{\small{Livebot}}                          & \multicolumn{4}{c}{\small{VideoIC}}               \\ 
\small{Model}   & \small{Br@1} & \small{Br@4} & \small{Bs@1}$^\downarrow$  &\small{Bs@4}$^\downarrow$  & \small{Br@1} & \small{Br@4} & \small{Bs@1}$^\downarrow$  &\small{Bs@4}$^\downarrow$ \\ \midrule
S2S    & 9.705&4.210 &88.90 &88.21 &7.056&5.698&84.23 &83.17    \\
F-RNN  ~\cite{2018LiveBot}  &3.549 &5.088 & 94.72 &88.97 &13.09 &9.896&93.81 &90.53    \\
U-Trans  ~\cite{2018LiveBot} &6.759 &2.566&92.19 &91.18 &15.07 &12.05&91.43 &90.80    \\
MML-CG   ~\cite{2020VideoIC}  &9.027 & 5.256 &91.21 &90.04     &  17.73 &13.66 &  91.17 &90.11                  \\    \hline
So-TVAE$_{3}$ &15.13&5.159&31.33&25.38                   &\textbf{23.15} &\textbf{16.57} &18.90 &16.13\\ 
So-TVAE$_{5}$ &\textbf{16.26}&\textbf{5.618}&29.24&24.30            &20.61 &14.44 &14.30 &12.17 \\
So-TVAE$_{10}$&14.57&4.806&\textbf{27.45}&\textbf{22.94}         &19.87 &14.20 &\textbf{11.88} &\textbf{10.22}\\\bottomrule 
\end{tabular}}
    \vspace{-0.37cm}  
\end{table*}

\subsection{Diversity Analysis}
Based on the diversity evaluation protocol mentioned before, we evaluate the generation diversity and quality of our So-TVAE and previous models. To explore the impact of sentimental categories on generative diversity, we consider more fine-grained experiments with setting sentimental categories N to $3$, $5$ and $10$. 
The comparison results are shown in Tabel \ref{diversity}.
Clearly, our model obtains a significant improvement compared with the previous methods especially in diversity (\text{Bs@1} from $\textbf{88.90}$ to $\textbf{27.45}$), indicating that our model can effectively improve the generation diversity, further promoting the quality of generated comments. 
In addition, for fine-grained sentimental classification, our model achieves the best generation diversity with $N=10$, but the best generation quality with $N=5$ or $N=3$. 
   This is because with the refinement of sentimental classification, generation diversity simply increases.
However, due to the sample limitation and increased training difficulty, the model is difficult to ensure the quality of generated comments of all sentiments, which may lead to a decline of the average generation quality.  Thus, the model achieves the best average quality when the sentimental category corresponds to the sample category of the dataset.

\begin{figure}
  \setlength{\abovecaptionskip}{0.2cm} 
    \vspace{-0.05cm}  
\begin{minipage}{.48\linewidth}
\centering
\includegraphics[width=5cm]{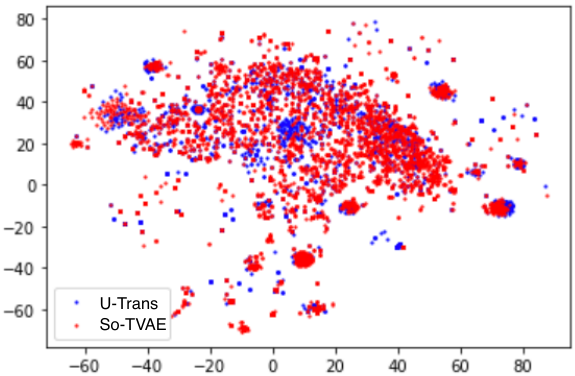}
\caption{Visualization of generated comments representation in 2D space with using \text{t-SNE}. We plot the scatter diagrams for $8,000$ samples.}
\label{visualization}
\end{minipage}%
\hfill
\begin{minipage}{.45\linewidth}
\captionof{table}{Comparison of average points distance after dimension reduction. Random: the ideal result for reference, corresponding to uniform diversity.}
\label{visualization_tab}
\centering
\renewcommand\arraystretch{1.1}
\resizebox{1\linewidth}{!}{
\begin{tabular}{@{}l|ll|l@{}}
\toprule
  \small{Model}  &  \small{So-TVAE}  &  \small{U-Trans}   & \small{Random} \\ \midrule
Distance &\textbf{58.56} &52.12 &72.63\\
 Normalized Distance &\textbf{0.806} &0.718 &1.00\\\bottomrule
\end{tabular}}
\end{minipage}
    \vspace{-0.27cm}  
\end{figure}

Further, to demonstrate the effectiveness of diversity modeling, we use the t-distributed stochastic neighbor embedding (\text{t-SNE}) ~\cite{2008Visualizing} visualization technique to analyze the discriminability of generated comments. Specifically, we use the proposed model and baseline model \text{U-Trans} to generate comments for $8,000$ samples in the VideoIC test set, and extract a $768$-dimension representation vector for each generated comment using the pre-trained text encoder \text{BERT} ~\cite{2018BERT}. 
Figure \ref{visualization} shows the visualization results of comment representation distribution after vector dimension reduction using the \text{t-SNE} ~\cite{2008Visualizing} tool. Intuitively, we can observe the comment representation cluster produced by \text{So-TVAE} is a wider coverage compared with the baseline model, indicating that \text{So-TVAE} can generate richer content representation. 
In addition, we further obtain a more intuitive comparison by calculating the mean point distance of $8,000$ comments after dimension reduction, with the smaller distance corresponding to less diversity. 
The calculation of random distribution is used as the ideal result for reference, which is the average distance of $8,000$ points randomly generated in this region, representing uniform diversity.
The results shown in Tabel \ref{visualization_tab} further prove that the comments provided by our \text{So-TVAE} are more discriminative and diverse.

    \vspace{-0.15cm}  
\subsection{Controllability Analysis}
Considering the application scenarios of live video commenting, the model may be asked to generate comments of the required sentimental types. Therefore, we conduct experiments to verify the performance of the model while directly generating comments according to the sentimental types of target comments. The results are reported in Table \ref{Controllability}.
The results show that our So-TVAE exhibits excellent performance with the guidance of the required sentiment, indicating that our model can effectively generate comments of the required sentimental types. This innovation further improves the applicability of live video commenting task for social media.

    \vspace{-0.15cm}  
\subsection{Extensive Experiments for Cross-modal Automatic Commenting}
To further demonstrate the superiority of our model in the field of cross-modal automatic commenting, we carry out some expansion experiments on another similar task image news commenting ~\cite{2019Cross}. Image news commenting, which aims to generate a reasonable and fluent comment with respect to the news body and images, is a novel task proposed in response to the interaction in online news.

\begin{table*}[]
  \setlength{\abovecaptionskip}{0.2cm} 
    \vspace{-0.27cm}  
\caption{Measurement of comment generation controllability, where R@k, MR and MRR are short for Recall@k, Mean Rank and Mean Reciprocal Rank. $_{3, 5, 10}$ annotates the results when setting $N=3,5, 10$. $^{\dagger}$ annotates the results generated according to the corresponding sentiments. R@k is reported as percentage (\%). $^\downarrow$ indicates that the performance is better with a lower score.}
\label{Controllability}
\centering
\resizebox{0.8\linewidth}{!}{
\begin{tabular}{@{}l|lllll|lllll@{}}
\toprule
 & \multicolumn{5}{c}{\small{Livebot}}                          & \multicolumn{5}{c}{\small{VideoIC}}               \\ 
  \small{Model}      & \small{R@1} & \small{R@5} & \small{R@10} & \small{MR}$^\downarrow$ & \small{MRR}  & \small{R@1} & \small{R@5} & \small{R@10} & \small{MR}$^\downarrow$ & \small{MRR}  \\
  \midrule
So-TVAE$^{\dagger} _{3}$& 27.47&53.57&{67.68}&10.17&0.403                &28.21 &55.33 & 69.95 &9.455 & 0.415\\
So-TVAE$^{\dagger} _{5}$&\textbf{27.83}&\textbf{53.66}&\textbf{68.52}&\textbf{10.13}&\textbf{0.405} &28.45 & 55.67 & 70.71 & 9.297 & 0.417 \\
So-TVAE$^{\dagger} _{10}$&25.08&{53.26}&{67.83}&{10.18}&0.388                     &\textbf{29.53} & \textbf{56.53} & \textbf{71.21} & \textbf{9.121} & \textbf{0.426}   \\\bottomrule
\end{tabular}}
    \vspace{-0.35cm}  
\end{table*}

\begin{table*}[]
  \setlength{\abovecaptionskip}{0.15cm} 
    \vspace{-0.05cm}  
\caption{Performance comparisons with the state-of-the-art methods for image news commenting. B@K, Br@k and Bs@k are short for BLEU@K, {BLEU}$_{ref}$ and {BLEU}$_{self}$, which are all reported as percentage (\%). $^\downarrow$ indicates that the performance is better with a lower score. $^{\dagger}$ annotates the results we reproduce on the baseline model. $_{3, 5, 10}$ annotates the results when setting $N=3, 5,10$. } 
\label{CMAC}
\centering
\resizebox{0.75\linewidth}{!}{
\begin{tabular}{@{}l|ll|ll|llll@{}}
\toprule
  \small{Model}      & \small{B@1} & \small{Rouge-L}  & \small{CIDEr} & \small{Meteor}  & \small{Br@1}  & \small{Br@4} & \small{Bs@1}$^\downarrow$ & \small{Bs@4}$^\downarrow$  \\  \midrule 
  CMAC-S2S ~\cite{2019Cross}&7.1 &9.1 &--&--&--&--&--&--\\
  CMAC-Trans ~\cite{2019Cross}& 7.7 &9.4  &—&—&—&—&—&—\\\hline
    CMAC-S2S$^{\dagger}$ ~\cite{2019Cross}& 7.12 &8.41 &3.92& 3.19    & 26.03 &3.128 &98.81 &97.72 \\
  CMAC-Trans$^{\dagger}$ ~\cite{2019Cross}&7.34 & 9.06  &3.41 &3.43    &29.08 &4.588 &97.69 &96.95\ \\
  \hline
So-TVAE$_{3}$&\textbf{8.47}&\textbf{9.77}  & {7.69} &\textbf{4.46}     &\textbf{40.30} &\textbf{4.827} &23.04 &17.87\\
So-TVAE$_{5}$ &8.38 &{9.56} &7.41 &4.19      &36.15 &4.410 &21.02 &17.01\\
So-TVAE$_{10}$ &7.96 &9.34 &\textbf{9.14} &3.74        &35.12. &4.376 &\textbf{20.66} &\textbf{15.18}\\\bottomrule
\end{tabular}}
    \vspace{-0.35cm}  
\end{table*}

 \begin{itemize}
  \item \textbf{Dataset}: The Cross-Modal Comments Dataset which is collected from a popular Chinese news website called Netease News \footnote{\url{https://news.163.com}}, consists of 24,134 news and 930,656 related comments. Each news sample contains an average of $5.80$ images, $54.77$ words of news body and $12.19$ words of news comments.

  \item \textbf{Metrics}: We first  adopt the evaluation protocol of ~\cite{2019Cross} for comparison. Then we further report the common text evaluation metrics \text{METEOR}~\cite{2005METEOR} and \text{CIDEr}~\cite{2015CIDEr}, and our proposed diversity metrics.
   For the diversity evaluation metrics, the baseline models are paired with beam search to generate multiple comments to compare. 

  \item \textbf{Details}: We use the following two baseline methods: CMAC-S2S and CMAC-Trans ~\cite{2019Cross}. For our So-TVAE, we conduct experiments based on different sentimental categories for comparison. The network parameters, optimizer and learning rate are fully similar to the above settings on dataset Livebot.
  \end{itemize}

Table \ref{CMAC} summarizes the performance comparisons between the state-of-the-art models and our So-TVAE. In general, our So-TVAE outperforms the baseline models in all metrics and achieves the new SOTA. 
Similarly, So-TVAE brings significant improvements in diversity metrics (\text{Br@1} from $\textbf{29.08}$ to $\textbf{40.30}$ and \text{Bs@1} from $\textbf{97.69}$ to $\textbf{20.66}$), again demonstrating the superiority of our diversity encoding. 
Moreover, the improvement of our model in the \text{CIDEr} score reached $\textbf{133.2\%}$ compared with the baseline models, as the CIDEr metric is more similar to the human judging mechanism, indicating that our model can generate more human-like comments for the image news commenting task.

\begin{figure*}[!t]
  \vspace{-0.4cm}  
    \setlength{\abovecaptionskip}{0.cm} 
  \setlength{\belowcaptionskip}{-0.29cm}
\centering
\includegraphics[width=5.2in]{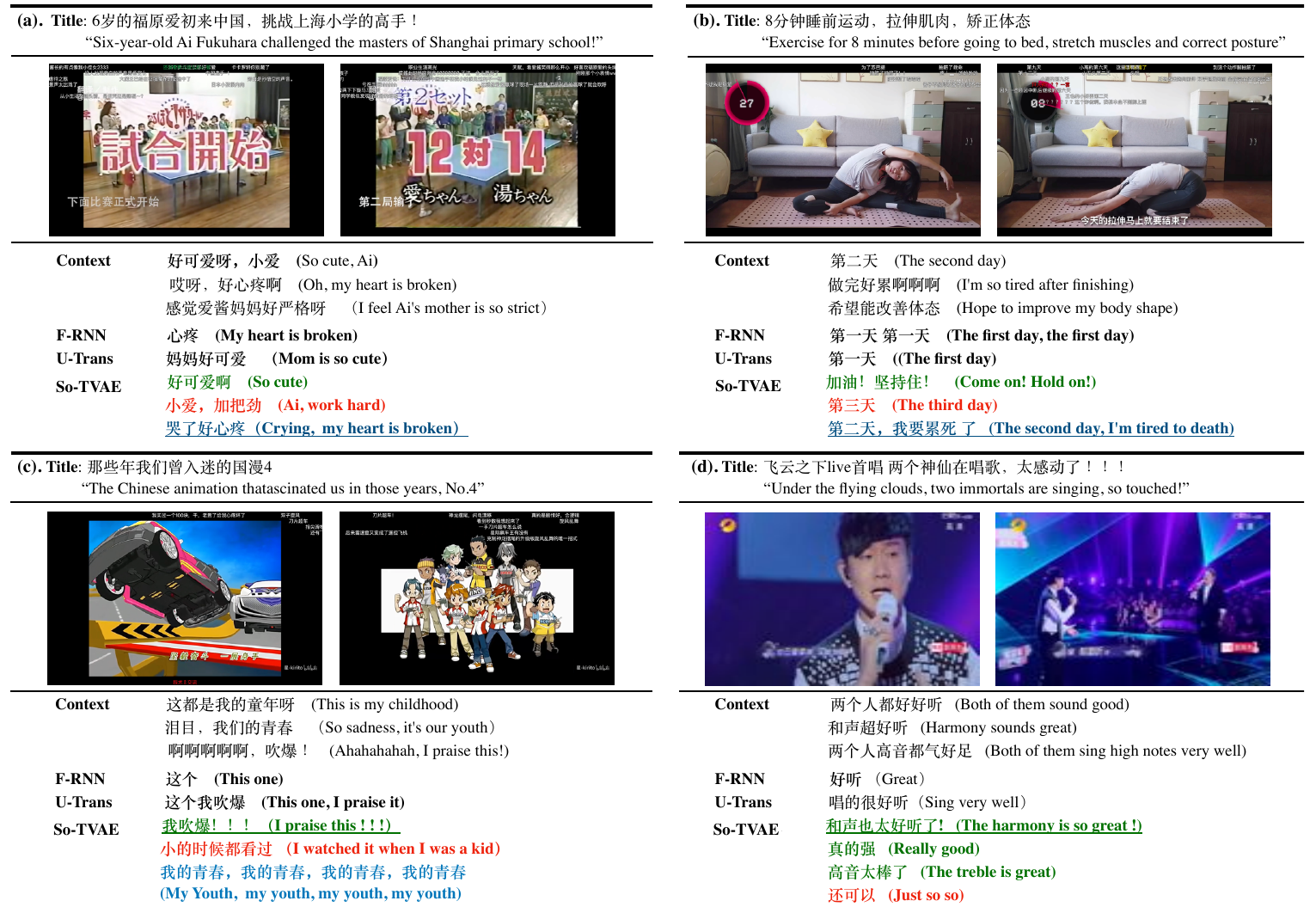}
\caption{Examples of generated comments by the baseline and So-TVAE for live video commenting. Green: positive comments. Red: neutral comments. Blue: negative comments. Context: human posted comments in context. The underline marks the generated comments of predicted sentimental type.}
\label{fig_3}
\end{figure*}

  \begin{figure*}[!t]
    \vspace{-0.05cm}  
      \setlength{\abovecaptionskip}{0.cm} 
  \setlength{\belowcaptionskip}{-0.29cm}
\centering
\includegraphics[width=3.8in]{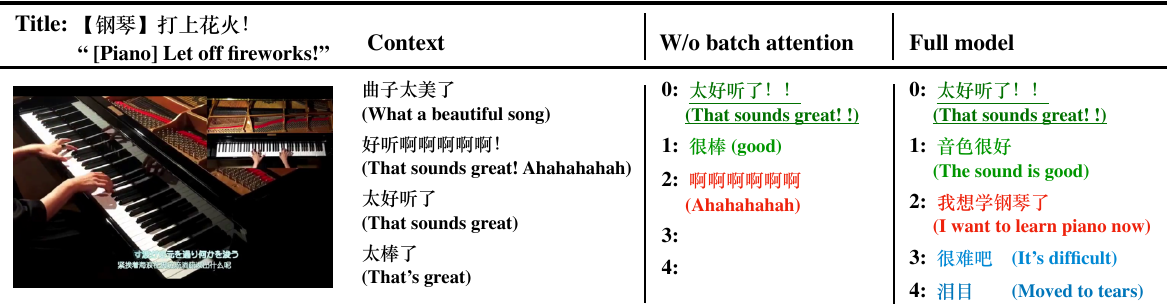}
\caption{{Example of generated comments by the model variants with or without batch attention for $N=5$. $0,1,2,3$, and $4$ respectively correspond to positive, slightly positive, neutral, slightly negative, and negative.}}
\label{batch_case}
\end{figure*}

  \begin{figure*}[!t]
    \vspace{-0.4cm}  
    \setlength{\abovecaptionskip}{0.cm} 
  \setlength{\belowcaptionskip}{-0.29cm}
\centering
\includegraphics[width=5.1in]{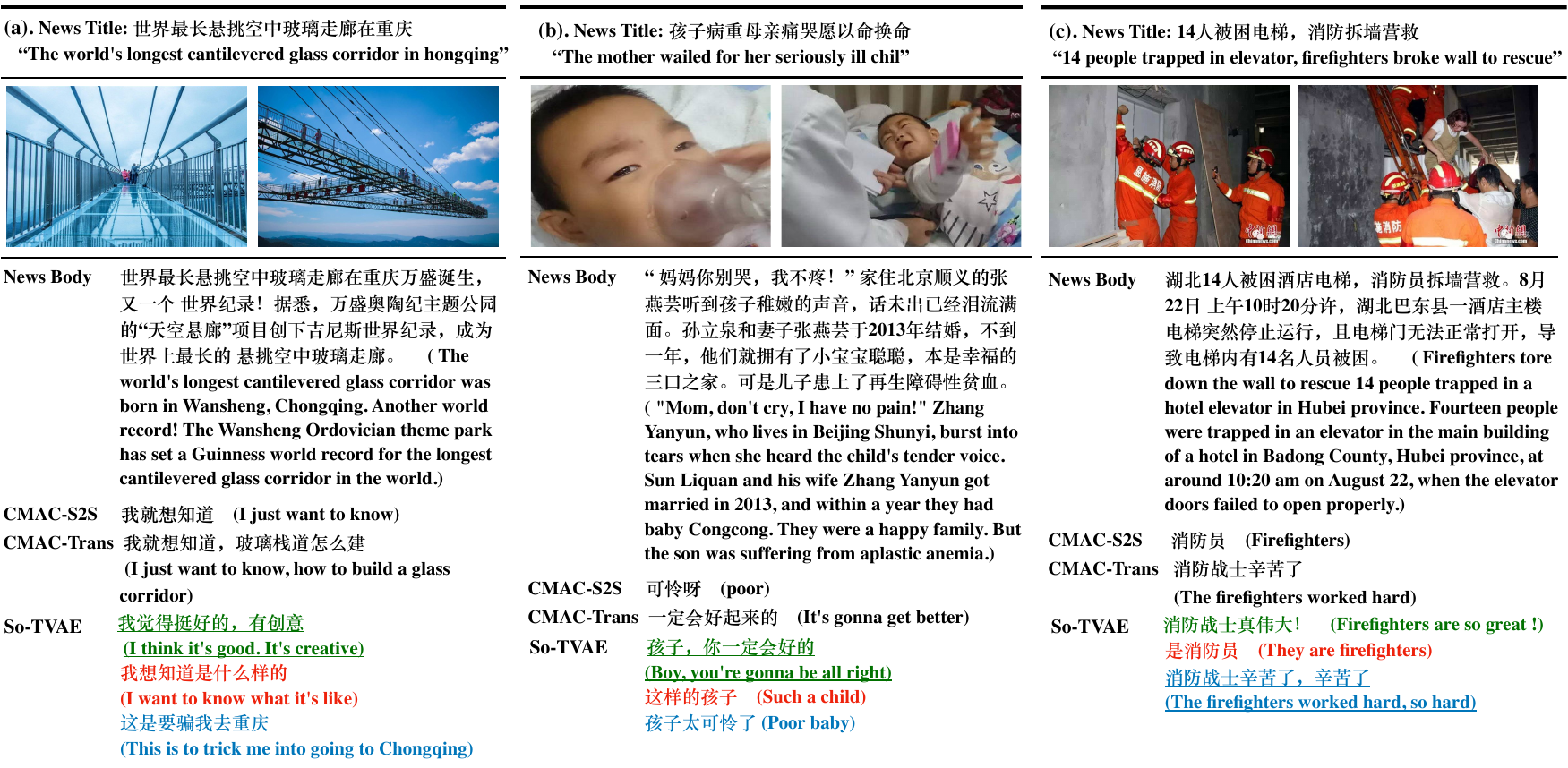}
\caption{Examples of generated comments by the baseline and So-TVAE for image news commenting. Green: positive comments. Red: neutral comments. Blue: negative comments.The underline marks the generated comments of predicted sentimental type.}
\label{CMAC_1}
    \vspace{-0.17cm}  
\end{figure*}

\subsection{Qualitative Analysis}
To provide a more direct qualitative comparison, we further visualize some representative results. 
{Firstly, to more intuitively exhibit the benefits of batch attention module for data imbalance problem, we visualize a generation case of the model variants with or without batch attention, with fine-grain sentimental category $N=5$. As shown in Figure \ref{batch_case}, in the context of a significant lack of negative and neutral comments, our So-TVAE with batch attention can generate diverse and related comments of all sentiments, indicating that the batch attention module based on sample relationship learning effectively introduces virtual samples to assist the learning of missing sentiments (sentiment type $4$ and $5$). }
Secondly, Figure \ref{fig_3} exhibits several live video commenting results of the baseline models and our So-TVAE, coupled with the video frames and contextual comments. We report the comments generated by So-TVAE under the guidance of three different sentimental types: positive (green), neutral (red), and negative (blue),  where the predicted sentimental type is underlined. 
Generally, compared with the baseline model, our So-TVAE produces more accurate and detailed comments in the single generation case. Taking the image of Figure \ref{fig_3} (b) as an example, our model generates a more detailed and accurate phrase ``tired",  while the baseline models only predict a high-frequency phrase ``the first day".
On the one hand, as we can see from Figure \ref{fig_3}, by choosing different sentimental types, the current model can successfully generate comments of different sentiments for a single video frame. 
On the other hand, after selecting the same sentimental type, our model can generate comments of different semantics with the same sentiment, such as ``Really good" and ``The harmony is so great'' in Part (d) of Figure \ref{fig_3}. 
These results collectively verify the effectiveness of the sentimental diversity module and the semantic diversity module.

We also investigate the generation effect of image news commenting with qualitative results. As is shown in Figure \ref{CMAC_1}, So-TVAE tends to generate more detailed and relevant comments than the baseline model, while the baseline models occasionally generate meaningless general comments. In addition, our model successfully generates relevant comments of different sentimental types, such as ``Boy, you're gonna be all right'' and ``Poor baby" in part (b), indicating that the one-to-many modeling can make better use of the diversity information in target comments.

In general, compared with the previous models, the sentiment-oriented multi-semantic comment generation method is more in line with human thinking habits, resulting in more humanized comments with the language characteristics of multiple perspectives and multiple sentiments.

\section{Limitation Analysis}
\label{Limitation}
{In the experiment, we observe some limitations of the model. 
Firstly, although our proposed batch attention module has improved the issue of data imbalance, in some samples, due to the significant lack of comments of corresponding sentiment, the model still finds it difficult to learn this type of comment information and tends to generate an empty sentence, such as the case of neutral comment in Part (d) of Figure \ref{fig_3}.  
Secondly, in a more complex chatting scenario, the model learns knowledge, but may still generate incorrect comments, such as the comment "My youth, my youth, my youth, my youth" in Part (c) of Figure \ref{fig_3}. 
Finally, although our model has effectively implemented the multi-modal information encoding for the ALVC task, the rich multi-modal information in live videos has not been fully utilized, such as audio information, captions information on video frames which can be obtained through optical character recognition, and so on.
Therefore, further solving the issue of data imbalance, enhancing the learning of hard samples, and fully utilizing the multi-modal information, all are feasible developing directions of the ALVC tasks.}

\section{Conclusion}
In this paper, we propose a Sentiment-oriented Transformer-based Variational Autoencoder (So-TVAE) for the ALVC task which can achieve diverse video commenting with multiple sentiments and multiple semantics. 
Specifically, we propose a sentiment-oriented diversity encoder to model the rich semantic diversity information and the sentimental diversity information in the target comments. To alleviate the mode imbalance problem, we propose a novel sentiment-oriented random mask mechanism to ensure the quality and diversity of the generation model. 
In addition, we also propose a batch attention module to model the sample relationship, which is useful to address the problem of missing sentimental samples caused by data imbalance.
Extensive experiments verify the effectiveness of our proposed So-TVAE in both quality and diversity.  
 In the future, our work can inspire other approaches to explore the sentiment-oriented live video commenting task and further encourage human-interacted comment generation.

\begin{acks}
\end{acks}

\bibliographystyle{ACM-Reference-Format}
\bibliography{sample-base}

\clearpage
\appendix

\section{Reparameterization Trick}
\label{repar}
{
We adopt the reparameterization trick ~\cite{2017Diverse, 2019Dispersed} to solve the gradient discontinuity problem in Gaussian mixture sampling. 
In the sentiment-oriented diversity encoder module, our model defines the Gaussian distribution with the mean vectors and the variances vector, then obtains the latent vector through Gaussian distribution sampling. 
As the non-differentiability of the sampling process leads to gradient discontinuity,  we adopt the reparameterization trick to enable the training and updating of the parameters of the Gaussian distribution.}

{
The core concept is to separate the random variable sampling process from the network, to enable the gradient directly propagated to the network parameters. 
Specifically, based on the conversion principle between a general Gaussian distribution and a standard normal distribution, i.e.:
 \begin{align}
\mathcal{N}(\mu, \sigma^{2}) \sim \mu+ \sigma \mathcal{N}(0,1),
\end{align}%
reparameterization transfers the randomness of the Gaussian sampling process to an independent random variable $\boldsymbol{\epsilon}$. Taking the Gaussian mixture sampling process of equation 5 as an example, a simple and effective reparameterization function can be set as:
 \begin{align}
p(\textbf{z}\mid\textbf{s})=\sum_{j=1}^N s_j (\mu_j+\sigma_j \epsilon_j), \ \ \  \epsilon_j  \sim \mathcal{N}(0,1)
\end{align}%
In this way, $\textbf{z}$ still satisfies the Gaussian mixture distribution with the mean $\boldsymbol{\mu}$ and the variance $\boldsymbol{\sigma}^2$, and the gradient can naturally propagate to the mean and variance, without being affected by the sampling process.}

\section{Datasets}
\label{data}
In this section, we provide a detailed explanation of the two live video commenting datasets Livebot ~\cite{2018LiveBot} and VideoIC ~\cite{2020VideoIC}, which  are all collected from a popular online video website, \text{BiliBili}.
\begin{itemize}
\item $\textbf{Livebot}$: 
Livebot consists of $2,361$ videos with a total duration of $114$ hours and $895,929$ comments with an average words count of $9.08$, forming $727,406$ samples. The dataset is split into $716,607$, $7,031$ and $3,768$ samples in the traning, testing and validating sets, respectively. 
\item $\textbf{VideoIC}$:
VideoIC is a large-scale dataset which consists of $4,951$ videos with a total duration of $557$ hours and $5,330,393$ comments with an average words count of $9.14$, forming $3,364,444$ samples. The dataset is split into $3,063,031$, $149,021$ and $152,392$ samples in the traning, testing and validating sets, respectively. 
\end{itemize}

\section{Co-attention Module}
\label{coa1}

\subsection{Definition}

The co-attention module adopts $L$ Transformer encoding layers follow by a weighting layer to achieve multi-modal interaction.
\begin{align}
 \hat{\textbf{X}},\hat{\textbf{Y}}  = \text{Co-Attention}(  \textbf{X}, \textbf{Y}), \ \  \ \ \ \ \ \ \ \ \ \ \ \ \ \
\end{align}
Taking $X$, $Y$ as the input, and the detailed implementation is:
\begin{align}
\textbf{X}^{l}&=\text{Transformer}_{x}^{l}(\textbf{X}^{l-1},\textbf{X}^{l-1},\textbf{X}^{l-1}), \ \ 
\textbf{Y}^{l}=\text{Transformer}_{y}^{l}(\textbf{Y}^{l-1}, \textbf{X}^{l},\textbf{X}^{l}).\\
 \hat{\textbf{X}} &=\sum_{i=1}^{k} \alpha_{i}X_i^{L},  \ \  for \ \ \lbrace \alpha_i \rbrace _{1}^{k} =\text{softmax}( \text{MLP}(\textbf{X}^{L}) ), \label{1}
\end{align}
where $\textbf{X}^{0}=\textbf{X}$ and  $\textbf{Y}^{0}=\textbf{Y}$. $\hat{\textbf{Y}}$ is obtained with a similar processing as Equation \ref{1}.

\subsection{More Ablation Results for Co-attention}

To further explore the structural rationality of co-attention module, we compared the following variants: 
(1) the models with different number L of Transformer encoding layers cascaded in depth.
(2) CoA1: the model with co-attention acting on $(\textbf{V}_\textbf{I}, \textbf{V}_\textbf{e})$.
(3) CoA2: the model with two co-attention acting on $(\textbf{V}_\textbf{e}, \textbf{V}_\textbf{I})$ and $(\textbf{V}_\textbf{I}, \textbf{V}_\textbf{e})$ respectively.
The results shown in Table \ref{COA} verified the effectiveness of the co-attention module for the full model.

\begin{table}[t]
\centering
\caption{More ablation results for co-attention module.}
\label{COA}
\renewcommand\arraystretch{1.1}
\resizebox{0.95\linewidth}{!}{
\begin{tabular}{l|lllll |lllll}
\toprule  
 & \multicolumn{7}{c}{\bf{Livebot}}                          & \multicolumn{3}{c}{\bf{VideoIC}}               \\ 
 Variant &  \bf R@1 & \bf R@5 &\bf R@10 &\bf MR$^\downarrow$ &\bf MRR &\bf R@1 &\bf R@5 &\bf R@10 &\bf MR$^\downarrow$ &\bf MRR 
\\ \hline 
L=2    & 24.10   &    48.86 &  63.35    & 11.90   & 0.366 &         {27.99} & {53.98} & {67.29} & {9.900} & {0.402}   \\
 L=4    &   25.58  &    50.12 &   65.40   &   11.20 &  {0.384}  &        29.25 & 55.17 & 69.77 & 9.530 &0.420    \\
  CoA1   &   24.73& 50.33& 64.95& 11.19& 0.379  &       27.33 & 53.29 & 68.15 &10.22 & 0.402     \\ 
  CoA2      & 22.72    &  48.89   &   63.56   &  11.88  &  0.358      &   22.23  &  49.19   &   64.67   &  11.23  &    0.356 \\
 \hline 
 So-TVAE      &\textbf{25.88} & \textbf{50.64} & \textbf{65.68} & \textbf{11.10} & \textbf{0.384}     &\textbf{29.58} & \textbf{55.35} & \textbf{69.67} & \textbf{9.538} &\textbf{0.413} \\
\bottomrule 
\end{tabular}}
\end{table}

\section{More Ablation Results for Loss Function}
\label{loss}
{
We adopt the control variable method to find a suitable setting of hyper-parameters. First fixing the hyper-parameter $\gamma=1$, by changing $\beta$ with different values $0.5, 1, 1.5$, the results indicate that the model achieved better performance with a higher $\beta$. Thus, through fine-grained hyper-parameter settings with $\beta=2, 2.5$, we can obtain a suitable setting of $\beta=2$. Then, by fixing $\beta$ and following the same steps, we can obtain a suitable setting for hyper-parameters of $\beta=2$ and $\gamma=0.3$. This conclusion holds for both datasets.}

\begin{figure*}[!t]
\centering
\includegraphics[width=5.5in]{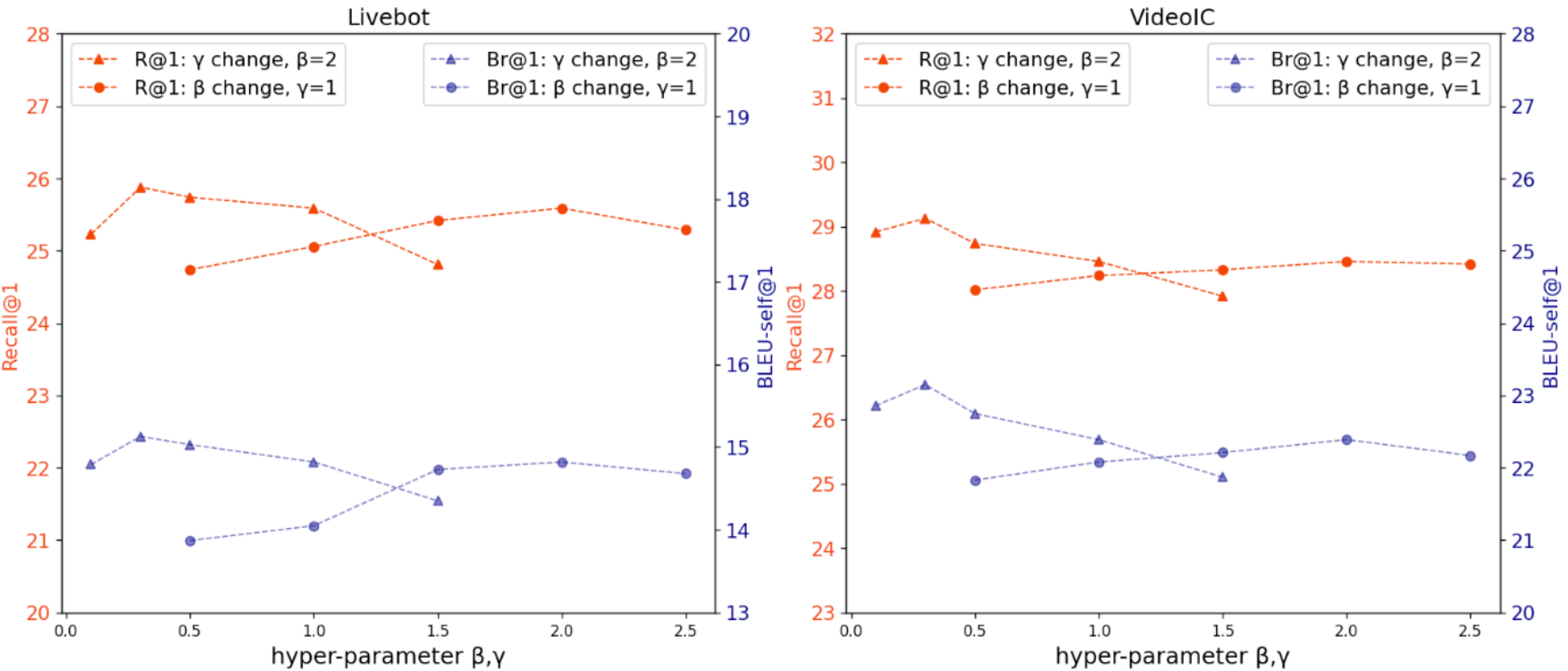}
\caption{The variation curves of model performance with respect to hyper-parameters  $\beta$ and $\gamma$ in metrics Recall@1 and $BLEU_{self}@1$.}
\label{lo}
\end{figure*}

{
A more intuitive comparison for the ablation studies of the hyper-parameters  $\beta$ and $\gamma$ in the loss function is provided, to illustrate the trend of model performance relative to hyper-parameters.
As shown in Figure \ref{lo}, obviously, our model tends to have a better performance with the higher $\beta$ and lower $\gamma$, and achieves the best performance with $\beta=2$ and $\gamma=0.3$.}


\section{Batch Attention module}
\label{Batch_add}

\subsection{More comparison with the BatchFormer}
\begin{figure*}[!t]
\centering
\includegraphics[width=5in]{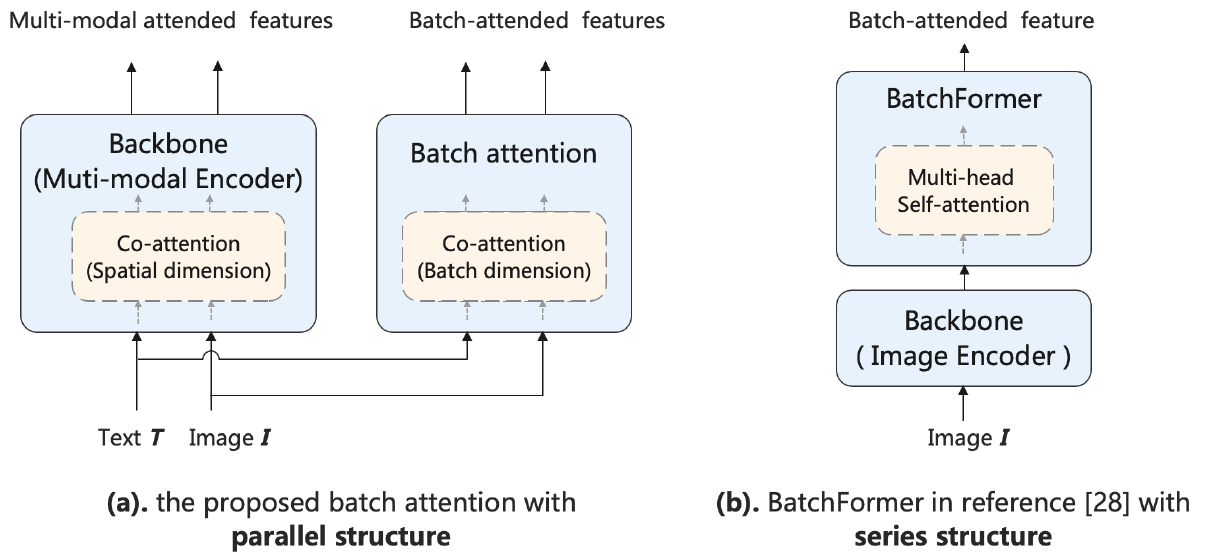}
\caption{A structural comparison diagram between our batch attention module and the BatchFormer module.}
\label{batch_structural}
\end{figure*}

{
In this section, we provide a detailed comparison of the proposed batch attention with the BatchFormer ~\cite{2022BatchFormer} module, in terms of the structural comparison diagram and ablation studies. Figure \ref{batch_structural} exhibits a structural comparison diagram between two modules, which more intuitively demonstrates the differences of batch attention in multi-modal input, algorithm structure, and attention mechanism. }

\begin{table*}[]
\caption{More ablation studies on the improvements of our batch attention module compared with the BatchFormer module, in algorithm structure (BatchFormer \& parallel structure) and attention mechanism (BatchFormer \& co-attention).}
\label{batch_al}
\centering
\renewcommand\arraystretch{1.1}
\resizebox{1\linewidth}{!}{
\begin{tabular}{@{}l|l|l|lllllll@{}}
\toprule

\small{Dataset  }                 &  \small{Module}               & \small{Variant }    & \small{R@1} & \small{R@5} & \small{R@10} & \small{MR}$^\downarrow$ & \small{MRR} & \small{Br@1} & \small{Bs@1}$^\downarrow$ \\ \midrule
\multirow{5}{*}{Livebot} 
                          & \multirow{3}{*}{Batch-attention}  & BatchFormer  & 22.93 &47.64 &61.94 &12.33 &0.354& 12.66 & 31.55      \\
                          &                              & BatchFormer \& co-attention     &24.54	&48.75	&63.29	&11.73	&0.366	&14.84	&31.46   \\ 
                          &                              & BatchFormer \&  parallel structure   &24.29&	48.37	&63.13	&11.57&	0.372&	14.71	&31.42     \\  \cline{2-10} 
    
                          & Full model                   & So-TVAE   &\textbf{25.88} &\textbf{50.64} & \textbf{65.68} & \textbf{11.10} &\textbf{0.384} & \textbf{15.13} &\textbf{31.33}          \\ \hline \hline

\multirow{5}{*}{VideoIC} 
                          & \multirow{3}{*}{Batch-attention}    & BatchFormer &29.00& 53.42 & 67.43 & 10.55 & 0.412 &22.75 &20.35 \\ 
                          &                              & BatchFormer \& co-attention    &29.04	&54.28&	68.95	&10.23	&0.416&	23.03&	19.42     \\ 
                          &                              & BatchFormer \&  parallel structure    &29.02	&54.07	&68.39&	10.07&	0.412&	22.84 &	19.68    \\  \cline{2-10}

                          & Full model                   & So-TVAE        & \textbf{29.13} & \textbf{54.91} & \textbf{69.37} & \textbf{9.712} & \textbf{0.419}  &\textbf{23.15} &\textbf{18.90}    \\ \bottomrule

\end{tabular}}
\end{table*}

{More ablation studies on the improvements of our batch attention module compared with the BatchFormer module are also conducted. As shown in Table \ref{batch_al}, the improvements in algorithm structure (BatchFormer \& parallel structure) and attention mechanism (BatchFormer \& co-attention) both contribute to the model performance, validating the benefits of our improvements in structure and attention mechanism for video commenting task. 
Furthermore, our proposed batch attention shows an average improvement of $9.12\%$ in all metrics compared with BatchFormer, proving the superiority of our proposed batch attention in video commenting tasks, which effectively combines the multi-modal interaction and batch interaction.}

\subsection{More Case Studies}
\begin{figure*}[!t]
\centering
\includegraphics[width=4.5in]{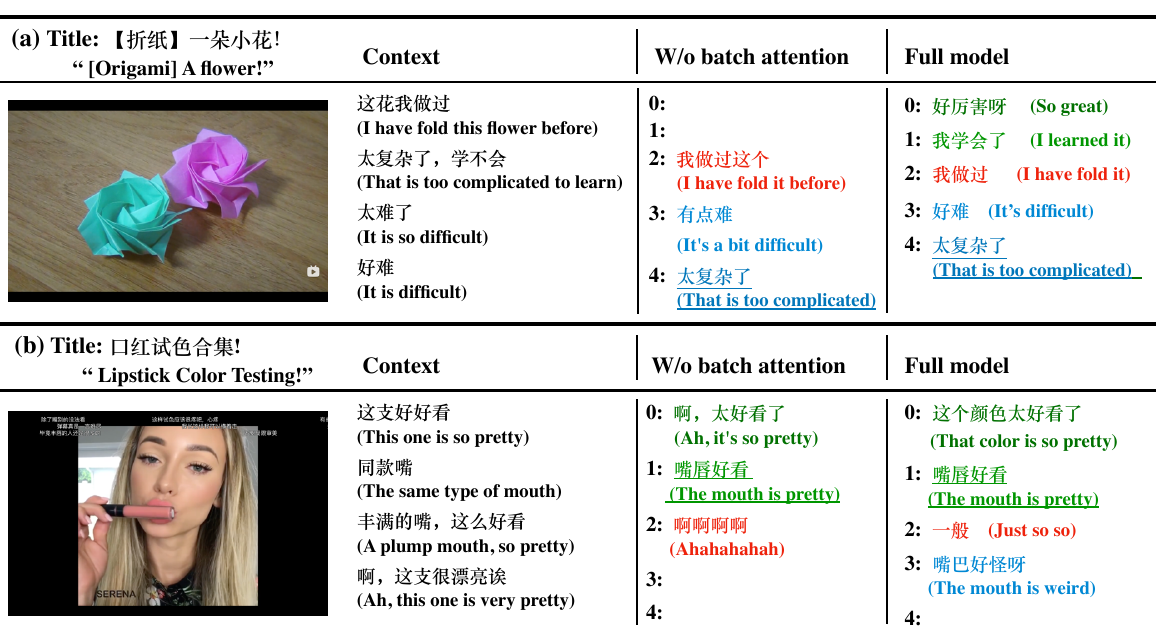}
\caption{More examples of generated comments by the model variants with or without batch attention for $N=5$. $0,1,2,3,4$ respectively correspond to positive, slightly positive, neutral, slightly negative, and negative.}
\label{batch_case2}
\end{figure*}

{ Besides, more generation cases of the model variants with or without batch attention are provided, with a fine-grain sentimental category $N = 5$. All the results in Figure \ref{batch_case2} demonstrate that our  So-TVAE with batch attention module effectively alleviates the data imbalance problem, yielding diverse comments even in the context of missing sentimental samples.
}

\end{document}